%% file: main.tex
\documentclass[11pt]{article}

\usepackage[margin=1.0in]{geometry}

\usepackage{amsmath}
\usepackage{graphicx}
\usepackage{color}
\usepackage{multirow}

\usepackage{rotating}
\usepackage{bbm}
\usepackage{latexsym}
\usepackage{times}
\usepackage{epsfig}
\usepackage{graphicx}
\usepackage{amsmath}
\usepackage{amssymb}

\usepackage{amssymb}
\usepackage{soul}
\usepackage{capt-of}
\usepackage{adjustbox}
\usepackage[utf8]{inputenc} 
\usepackage{latexsym}
\usepackage{amsthm}
\usepackage{cite}
\usepackage[font=small,labelfont=bf]{caption} 
\usepackage[subrefformat=parens]{subcaption}
\usepackage{epsfig}
\usepackage{subcaption}
\usepackage[pagebackref=true,breaklinks=true,colorlinks,bookmarks=false]{hyperref}
\usepackage{tabularx} 
\usepackage{threeparttable}

\usepackage{algorithm}
\usepackage{algcompatible}

\usepackage{bm}
\allowdisplaybreaks
\newcolumntype{L}{>{\raggedright\arraybackslash}m{4.25cm}}
\newcolumntype{C}{>{\centering\arraybackslash}m{2cm}}
\newcolumntype{N}{>{\centering\arraybackslash}m{2.5cm}}

\newcommand{\captionfonts}{\normalsize}

\makeatletter  
\long\def\@makecaption#1#2{%
  \vskip\abovecaptionskip
  \sbox\@tempboxa{{\captionfonts #1: #2}}%
  \ifdim \wd\@tempboxa >\hsize
    {\captionfonts #1: #2\par}
  \else
    \hbox to\hsize{\hfil\box\@tempboxa\hfil}%
  \fi
  \vskip\belowcaptionskip}
\makeatother   

\newtheorem{theorem}{Theorem}
\newtheorem{definition}{Definition}

\newtheorem{lemma}{Lemma}

\newcommand{\real}{{\mathbb{R}}}

\newcommand{\argmin}{\operatorname{argmin}}

\newcommand{\argmax}{\mathrm{argmax}}

\newcommand{\E}[1]{\underset{#1}{\mathbb{E}}}
\newcommand{\der}{\mathrm{d}}
\newcommand{\JSD}{\mathrm{JSD}}

\newcommand{\KL}{\mathrm{KL}}

\newcommand{\D}{\mathrm{D}}
\newcommand{\JR}{\mathrm{JR}}

\newcommand{\X}{\mathbb{X}}
\newcommand{\renyi}{R\'{e}nyi}
\newcommand{\PV}{\vert \chi \vert^k}

\usepackage{authblk}

\begin{document}

    \title{\bf Least $\bm{k}$th-Order and \renyi \ Generative Adversarial Networks}

\bigskip

\author[*]{\normalsize Himesh Bhatia}
\author[**]{\normalsize William Paul}
\author[*]{\normalsize Fady Alajaji}
\author[*]{\normalsize Bahman Gharesifard}
\author[**]{\normalsize Philippe Burlina}

\bigskip

\affil[*]{\small Department of Mathematics and Statistics, Queen's University, Kingston, ON K7L 3N6, Canada 
        (\texttt{himesh.bhatia@queensu.ca}, \texttt{fa@queensu.ca}, \texttt{bahman.gharesifard@queensu.ca}).}

\affil[**]{\small The Johns Hopkins University, Applied Physics Laboratory, Laurel, MD 20723, USA
        (\texttt{william.paul@jhuapl.edu}, \texttt{philippe.burlina@jhuapl.edu})}

\newcommand\blfootnote[1]{%
  \begingroup
  \renewcommand\thefootnote{}\footnote{#1}%
  \addtocounter{footnote}{-1}%
  \endgroup
}


\date{}
\maketitle


\begin{abstract} 
We investigate the use of parametrized families of information-theoretic measures to generalize the loss functions of generative adversarial networks (GANs) with the objective of improving performance.
A new generator loss function, called least $k$th-order GAN (L$k$GAN), is first introduced, generalizing the least squares GANs (LSGANs) by using a $k$th order absolute error distortion measure with $k \geq 1$ (which recovers the LSGAN loss function when $k=2$). 
It is shown that minimizing this generalized loss function under an (unconstrained) optimal discriminator is equivalent to minimizing the $k$th-order Pearson-Vajda divergence. 
Another novel GAN generator loss function is next proposed in terms of R\'{e}nyi cross-entropy functionals with order $\alpha >0$, $\alpha\neq 1$. 
It is demonstrated that this R\'{e}nyi-centric generalized loss function, which provably reduces to the original GAN loss function as $\alpha\to1$,
preserves the equilibrium point satisfied by the original GAN based on the Jensen-\renyi \ divergence, a natural extension of the Jensen-Shannon divergence. 
 
Experimental results indicate that the proposed loss functions, applied to the MNIST and CelebA datasets, under both DCGAN and StyleGAN architectures, confer performance benefits by virtue of the extra degrees 
of freedom provided by the parameters $k$ and $\alpha$, respectively. 
More specifically, experiments show improvements with regard to the quality of the generated images as measured by the Fr\'echet Inception Distance (FID) score and training stability.
While it was applied to GANs in this study, the proposed approach is generic and can be used in other applications of information theory to deep learning, e.g., the issues of fairness or  privacy in artificial intelligence.
\end{abstract}

\medskip

\begin{center} 
{\bf Keywords}
\end{center}
\vspace{-0.05in}
Deep learning, generative adversarial networks, \renyi \ cross-entropy, Jensen-\renyi \ divergence, Pearson-Vajda divergence.

\input{introduction.tex}

\input{divergences.tex}

\input{lkgan.tex}

\input{renyigan.tex}
\input{conclusion.tex}

\bibliographystyle{plain}
\bibliography{citations}

\appendix
\input{appendix.tex}

\end{document}

%% file: introduction.tex
\section{Introduction}
Generative models have garnered significant attention in recent years.
The main approaches for such models include generative adversarial networks (GANs) \cite{Goodfellow2014}, \cite{wasserstein2017}, \cite{Radford2015}, \cite{creswell2018generative}, autoencoders/variational autoencoders (e.g. VAEs) \cite{VAEs}, 
generative autoregressive models \cite{oord2016wavenet}, invertible flow based latent vector models \cite{kingma2018glow}, 
and hybrids of the above models \cite{grover2018flow}.
Compared to other approaches, GANs have generated the most interest (e.g., see surveys in~\cite{creswell2018generative}, \cite{wang2020generative}, \cite{wiatrak2020stabilizing}).
Generative models are used in reinforcement learning,  time series predictions, fairness and privacy in artificial intelligence (AI)~\cite{burlina2020addressing}, disentanglement~\cite{paul2020unsupervised}, and can also be trained in a semi-supervised manner, where labels and training examples are missing.
Furthermore, these models are designed to produce several different outputs that are equally acceptable~\cite{goodfellow_nips},~\cite{styleGAN}.

Approximation methods are needed in the case of VAEs and furthermore,
the approximate probability distribution is not guaranteed to converge to the true distribution~\cite{lsgan},~\cite{lsgan2},~\cite{wasserstein2017}.
In contrast, GANs optimize a loss function which is constructed using ideas from game and information theory. Notably, GANs can represent distributions that lie on low dimensional manifolds, which VAEs and estimating densities are unable to do \cite{wasserstein2017}.
Moreover, GANs do not rely on Markov chains and variational bounds are not necessarily needed~\cite{goodfellow_nips}.

\subsection{Prior work}
The original GANs \cite{Goodfellow2014} consist of a generative neural network
competing with a discriminative neural network in a min-max game.
Several variants of GANs have been studied and implemented. 
Deep convolutional GANs (DCGANs) use convolutional layers to learn higher dimensional dependencies 
that are inherent in complex datasets such as images~\cite{Radford2015}.  
Although DCGANs produced better results than other 
state-of-the-art generative models such as VAEs and autoregressive models, 
they can be difficult to train and can suffer from mode collapse~\cite{wasserstein2017, wiatrak2020stabilizing}.
Mode collapse occurs when the generator produces outputs that significantly lack in diversity (i.e. producing mostly one output with little variations), while the discriminator is able to tell apart real from fake
data perfectly during training.
Researchers have diligently attempted to fix the aforementioned issues.
For example, StyleGANs~\cite{styleGAN} change the architecture of the generative neural network to produce
realistic high resolution images, while Wasserstein GANs~\cite{wasserstein2017} address the problem of mode collapse by using the Wasserstein-1 distance as the loss function.

GANs have been applied to data privacy problems, where the goal is to hide certain features of user data to protect their privacy, but mask these features judiciously in order not to compromise other useful data \cite{gap}. Related problems of AI fairness can be addressed using this strategy.
GANs have also  been widely used for computer vision problems, such as generating fake images of handwritten numbers, or landscape paintings \cite{goodfellow_nips}. Thus the flexibility of GAN design allows for innovation and applicability to a wide range of data and use cases.

The use of information theory to study and improve neural networks is a relatively new yet promising direction of research; e.g., see~\cite{infomax, Nowozin2016fgan, principe10, whereInfo, infoGAN, renyiDNN, infoBottleneck0, infoBottleneck1, bridging, infoBottleneck} and the references therein.
While many GANs loss functions are based on the Jensen-Shannon divergence, there are other divergence measures and tools in information theory
that can be directly applied to the design of GANs.  
The family of loss functions that simplify down to $f$-divergences was thoroughly studied in
\cite{Nowozin2016fgan, convexDuality}, and~\cite{bridging2}.
Bridging the gap between maximum likelihood learning and 
GANs, especially those with loss functions that simplify down to $f$-divergences, 
has also been analyzed in \cite{bridging}.
Using the symmetric Kullback-Leibler (KL) divergence, researchers have also shown that a variant of VAEs
is connected to GANs \cite{svaes}.
InfoGANs use variational mutual information maximization with latent codes to achieve unsupervised
representation learning with considerable success~\cite{infoGAN}.

A new least squares loss function that simplifies down to the Pearson $\chi^2$ divergence
was examined in~\cite{lsgan}. Through experiments, it was illustrated that the resulting least squares GANs (LSGANs) are more stable than DCGANs.
The promising results of LSGANs and the fact that the Pearson-Vajda $\vert \chi \vert^k$
divergence of order $k \geq 1$ generalizes the Pearson $\chi^2$ divergence is one motivation for this study. 

The use of the \renyi \ divergence in the context of GANs is a recent development. 
\renyi \ used the simplest set of postulates that characterize Shannon's entropy and 
introduced his own entropy and divergence measures (parametrized by its order~$\alpha$) 
that generalize the Shannon entropy and the KL divergence, 
respectively~\cite{renyi}. 
Moreover, the original Jensen-\renyi \ divergence~\cite{jenRenyi} as well as the
identically named divergence~\cite{kluza19} used in this paper
are non-$f$-divergence generalizations of the Jensen-Shannon divergence. 
Traditionally, \renyi's entropy and divergence have had 
applications in a wide range of problems, including lossless data compression 
\cite{llc, lossless, fady0},
hypothesis testing~\cite{csiszar, cutoff04}, 
error probability~\cite{error}, and guessing~\cite{guess, mutualInfo}.
Recently, the \renyi \ divergence and its variants (including Sibson's mutual information) 
were used to bound the generalization error in learning algorithms~\cite{esposito20}, and to analyze deep neural networks (DNNs)~\cite{renyiDNN}, 
variational inference~\cite{varInf}, Bayesian neural networks~\cite{bnn}, and generalized learning vector quantization~\cite{vecQ}.
Furthermore Cumulant GANs proposed a new loss function using the cumulant generating function, which subsumes the \renyi, KL, and reverse KL divergences as special cases, and is experimentally shown to be more robust than Wasserstein GANs~\cite{cumulant_gan}.
Hence, generalizing the GANs loss function using \renyi \ divergence provides another motivation for this work.

\subsection{Contributions} 
The novel contributions of this work are described in what follows. We revisit the LSGAN and original GAN generator loss functions by considering more general parametrized classes of loss functions that subsume the original loss functions as a special case.  An important objective is to identify generalized loss functions that can be analytically minimized under an (unconstrained) optimal discriminator, with the minimum theoretically achieved when the generator’s distribution is the true dataset distribution. 

More specifically, we first introduce \textit{least $k$th-order GANs (L$k$GANs)} by using 
the $k$th-order absolute error loss function for the generator ($k \geq 1$). 
We prove that minimizing this loss function is equivalent 
to minimizing the $k$th-order Pearson-Vajda divergence \cite{nielsen}, which recovers the
Pearson $\chi^2$ divergence examined in \cite{lsgan} when $k = 2$.
The L$k$GANs' generator loss function also preserves the theoretical minimum of LSGANs' generator loss function, 
which is achieved when the 
generator's distribution is equal to the true distribution.   
L$k$GANs are implemented and compared with LSGANs on the CelebA \cite{celeba} dataset using the StyleGAN architecture. 
Experimentally, L$k$GANs are shown to outperform LSGANs in terms of generated image quality, as measured by the Fr\'echet inception distance (FID) score~\cite{Heusel}, and the rate at which they converge to meaningful results.
L$k$GANs are also observed to reduce the problem of mode collapse during training. 

We also revisit the original GAN's generator optimization problem via the novel use of \renyi \ information measures of order $\alpha$
(with $\alpha>0$ and $\alpha\neq 1$).
To this end, we consider the Jensen-\renyi \ divergence
as well as the differential \renyi \ cross-entropy, which is the continuous form of the \renyi \ cross-entropy studied in~\cite{renyicrossentropy}.
We show that the differential \renyi \ cross-entropy generalizes the Shannon differential cross-entropy and prove that it is monotonically decreasing in $\alpha$. 
With these \renyi \ measures in place, we propose a new GAN's generator loss function
expressed in terms of the negative sum of two
R\'{e}nyi cross-entropy functionals. 
We show that minimizing
this $\alpha$-parametrized loss function under an optimal discriminator results in
the minimization of the Jensen-\renyi \ divergence~\cite{kluza19}, which is a natural extension 
of the Jensen-Shannon divergence as it uses the R\'{e}nyi divergence instead
of the Kullback-Leibler (KL) divergence in its expression.\footnote{Note that 
this Jensen-\renyi \ divergence measure,
which reduces to the Jensen-Shannon divergence as $\alpha$ approaches~1, 
differs from an earlier namesake measure introduced in~\cite{jenRenyi, jenRenyi2}
and defined using the \renyi \ entropy.} 
We also prove that our 
generator loss function of order $\alpha$ converges to the original GAN loss function
in~\cite{Goodfellow2014} when $\alpha \to 1$.
Previously, the GANs loss function has been generalized using the $f$-divergence
measure~\cite{csiszar67, Nowozin2016fgan}.  
However, as the Jensen-\renyi \ divergence is not itself an $f$-divergence,
it can be interpreted as a non-$f$-divergence generalization of the Jensen-Shannon divergence.
We call the resulting GAN network \renyi GAN.

In a concurrent work~\cite{rgan}, a \renyi-based GAN is developed, called RGAN, by also using a ``\renyi \ cross-entropy'' to generalize the original GAN loss function and it is shown that the resulting loss function is stable in terms of its absolute condition number (calculated using functional derivatives).
Note that \renyi GAN is based on a {\em different} \renyi \ cross-entropy definition\footnote{Specifically in~\cite[Equations~((1.5) and~(1.6)]{rgan}, the \renyi \ cross-entropy between two distributions $p$ and $q$ is defined as the sum of the \renyi \ divergence between $p$ and $q$ and the \renyi \ entropy of $p$. This definition is indeed different from the one we adopt herein (see Definition~\ref{r-cross-entropy}), as the continuous analogue of the one introduced in~\cite{renyicrossentropy}.} from the one used in RGAN~\cite{rgan}.
Hence, unlike what it is claimed in~\cite{rgan} by referencing the preprint \cite{paper} of this paper,
the RGAN's loss function does not generalize the \renyi GAN loss functions presented herein.
Moreover unlike \renyi GAN, the RGAN loss function does not preserve the GANs theoretical result that the optimal generator for an optimal discriminator induces a probability distribution equal to the true dataset distribution.
Using a similar stability analysis to the one carried out for RGANs in~\cite{rgan}, we derive the absolute condition number
of our \renyi \ cross-entropy measure and 
show that the \renyi GAN's generator loss function is stable for $\alpha \geq 2$. This result complements the one derived in~\cite{rgan}, where stability of the RGAN loss function is shown for {\em sufficiently small} values of $\alpha$.

Finally, we implement the newly proposed \renyi GAN loss function using the DCGAN and StyleGAN architectures~\cite{styleGAN}.  
Our experiments use the MNIST~\cite{mnist} and CelebA~\cite{celeba} datasets and provide comparisons with the baseline DCGAN and StyleGAN systems.
Experiments show that the R\'{e}nyi-centric GAN systems perform as well as, or better, than their baseline counterparts in terms of visual quality
of the generated images (as measured by the FID score), 
particularly when spanning $\alpha$ over a range of values as it helps the avoidance of local minimums. 
We show that employing $L_1$ normalization with the \renyi \ generator loss function confers greater stability, 
quicker convergence, and better FID scores for both \renyi GANs and DCGANs.
Consistent stability and slightly improved FID scores are also noted when comparing R\'{e}nyiStyleGAN with StyleGAN.
We finally compare these GAN systems with the simplified gradient penalty \cite{mescheder2018}, 
showing that the \renyi-type systems provide substantial reductions in computational training time vis-a-vis the baselines, for similar levels of FID.
\footnote{Our codes, network architectures, and results can be found at https://github.com/renyigan-lkgan?tab=repositories.}

The rest of the paper is organized as follows.
In Section~\ref{sec:divMeasures}, we present the definitions of divergence measures and introduce a natural extension of the (differential) Shannon cross-entropy, the (differential) \renyi \ cross-entropy. 
In Section~\ref{sec:lkgan}, we analyze L$k$GANs, which generalize LSGANs, and provide experimental results. 
In Section~\ref{sec:renyigan}, we present theoretical results of \renyi GANs, 
which generalize GANs, and show experimental results comparing \renyi GANs with DCGANs and StyleGANs. 
Finally, we provide conclusions in Section~\ref{sec:conclusion}.

%% file: divergences.tex
\section{Divergence Measures}
\label{sec:divMeasures}
Divergence measures are used to quantify the dissimilarity between distributions.
We recall the definitions of the Kullback-Leibler divergence, the (differential) Shannon cross-entropy, and the Pearson-Vajda and \renyi \ divergences.
We also present the definition of the (differential) \renyi \ cross-entropy and examine some of its properties.
We also present the Jensen-\renyi \ divergence, which is a natural extension of the Jensen-Shannon divergence by virtue of being a mixture of two \renyi \ divergences.
This Jensen-\renyi \ divergence was recently introduced in~\cite{kluza19} for discrete distributions
and studied in the context of generalized (\renyi-type) $f$-divergences. It differs from the identically named divergence studied in \cite{jenRenyi} and~\cite{jenRenyi2}, an earlier extension of the Jensen-Shannon divergence consisting of the difference between the \renyi \ entropy of a mixture of multiple probability distributions and the mixture of the \renyi \ entropies of the individual distributions.
Other recent (but different) extensions of the Jensen-Shannon divergence can be found in~\cite{nielsen19} and the references therein.

Let $p$ and $q$ be two probability densities
with common support $\mathcal{R} \subset \real$
on the Lebesgue measurable space $(\real, \mathcal{B}(\real), \mu)$, and let
\begin{equation}\KL(p \Vert q) := \int_{\mathcal{R}} p \log \frac{p}{q} \, \der \mu 
\qquad \text{and} \qquad
h(p;q):=-\int_{\mathcal{R}} p \log q \, \der \mu \label{kl-ce}\end{equation}
denote the KL divergence and the differential Shannon cross-entropy
between $p$ and $q$, respectively, where both information measures are assumed to be finite.
In~\eqref{kl-ce} and throughout, we use the short form $\int_{\mathcal{R}} f \, \der \mu
:= \int_{x\in\mathcal{R}} f(x) \, \der \mu(x)$ for any measurable function $f$.
When  $p = q$ almost everywhere (a.e.), then $h(p;q)$ reduces to the differential 
Shannon entropy of $p$, denoted by $h(p)$.
We next describe the Pearson-Vajda divergence, which is itself an $f$-divergence \cite{nielsen}.
\begin{definition}
        The \textbf{Pearson-Vajda divergence of order $\bm{k}$}, $\PV(p \Vert q)$, 
        between $p$ and $q$, where $k\ge 1$, is given by
        \begin{eqnarray} 
                \PV (p \Vert q) := \int_{\mathcal{R}} \frac{\vert q - p \vert^k}{p^{k-1}} \der \mu.
        \end{eqnarray}
\end{definition}
Note that $\PV (p \Vert q) \ge 0$ with equality if and only if (iff) $p = q$ (a.e.).
Also when $k = 2$, $\PV (\cdot \Vert \cdot)$ reduces to the {\bf Pearson $\bm{\chi^2}$ divergence}:
\begin{eqnarray}
                \chi^2(p \Vert q) := \int_{\mathcal{R}} \frac{(q - p)^2}{p} \der \mu.
        \end{eqnarray}
\begin{definition} 
         The \textbf{\renyi \ divergence of order} $\bm{\alpha}$ between $p$ and $q$, where $\alpha > 0$, $\alpha \neq 1$, is given by
        \begin{eqnarray}
                \D_{\alpha}(p \Vert q) := \frac{1}{\alpha - 1} \log \left (
                                \int_{\mathcal{R}} p^{\alpha}q^{1 - \alpha}\der \mu
                        \right ).
        \end{eqnarray}
\end{definition}
Note that $\D_{\alpha} (p \Vert q) \ge 0$ with equality iff $p = q$ (a.e.).
Furthermore, if $\D_{\gamma}(p \Vert q) < \infty$ for some $\gamma >1$,
then, as shown in~\cite{renyiDivergence}, we have that 
\begin{eqnarray} \label{renyi-KL}
\lim_{\alpha \rightarrow 1} \D_{\alpha}(p \Vert q) &=& \KL(p \Vert q). 
\end{eqnarray}
For simplicity of analysis, we assume in what follows the 
finiteness of $\D_{\gamma}(\cdot \Vert \cdot)$
for some $\gamma >1$ so that convergence of~\eqref{renyi-KL} holds.
Being a function of an $f$-divergence, useful properties and bounds 
on the \renyi \ divergence can be elucidated from the study of $f$-divergences, 
see~\cite{sason18} and related references.
\begin{definition}\label{r-cross-entropy}
        The \textbf{differential \renyi \ cross-entropy of order} $\bm{\alpha}$ between
        $p$ and $q$,
        where $\alpha > 0$, $\alpha \neq 1$, is given by
        \begin{eqnarray}
                 h_{\alpha} (p;q) &:=& \frac{1}{1-\alpha} \log \left (
                                \int_{\mathcal{R}} p q^{\alpha - 1} \der \mu    
                        \right ). 
        \end{eqnarray}
\end{definition}
When $p = q$ (a.e.),  $h_{\alpha} (p;q)$ reduces to the {\bf \renyi \ entropy}, $h_{\alpha} (p):= \frac{1}{1-\alpha}\log \left (     \int_{\mathcal{R}} p^{\alpha} \der \mu \right )$.
We next show that $\lim_{\alpha \rightarrow 1} h_{\alpha} (p;q) = h(p;q)$ under some finiteness conditions.
\begin{theorem} 
\label{theorem:renyi-crossentropy-limit}
        If $h(p;q) < \infty$, then
        \[\lim_{\alpha \downarrow 1} h_{\alpha} (p;q) = h(p;q).\]
        Moreover, if $\E{A \sim p}\left( \frac{1}{q(A)} \right) < \infty$, then
        \[\lim_{\alpha \uparrow 1} h_{\alpha} (p;q) = h(p;q).\]
\end{theorem}
The proof of the theorem requires the following result.
\begin{lemma} \cite{renyiDivergence} \label{erwen}
        For any $x > \frac{1}{2}$,
        \begin{eqnarray*}
                (x - 1) \left ( 1 + \frac{1-x}{2} \right )\leq \log(x) \leq x - 1.
        \end{eqnarray*}
\end{lemma}
{\bf Proof of Theorem~\ref{theorem:renyi-crossentropy-limit}} 
Working in the measure space $(\mathcal{R}, \mathcal{B}(\mathcal{R}), \mu)$, where $\mathcal{B}(\mathcal{R})$ is the Borel
$\sigma$-algebra and $\mu$ is the Lebesgue measure on $\real$, we first note that
	by setting $$x_{\alpha} = \E{\mathbf{A} \sim p}(q(\mathbf{A})^{\alpha - 1}) = \int_{\mathcal{R}} p q^{\alpha - 1} \der \mu$$
        we have that
        $\lim_{\alpha \downarrow 1} x_{\alpha} = 1$. 
        Also, Lemma~\ref{erwen} yields that
        $$\lim_{\alpha \downarrow 1} 
        \frac{\log\left (x_{\alpha} \right)}{x_{\alpha} - 1} = 1.$$
We then can write
        \begin{eqnarray*}
 \lim_{\alpha \downarrow 1} h_{\alpha}(p;q) 
                &=& \lim_{\alpha \downarrow 1} \frac{1}{1 - \alpha} \log\left (x_{\alpha} \right) \\
                &=& \lim_{\alpha \downarrow 1} \frac{x_{\alpha} - 1}{1 - \alpha}
                        \frac{\log\left (x_{\alpha} \right)}{x_{\alpha} - 1} \\
                &=& \lim_{\alpha \downarrow 1} \left( \frac{x_{\alpha} - 1}{1 - \alpha} \right)
                    \lim_{\alpha \downarrow 1} \left( \frac{\log\left (x_{\alpha} \right)}{x_{\alpha} - 1} \right)  \\
		&=& \lim_{\alpha \downarrow 1} \frac{x_{\alpha} - 1}{1 - \alpha}
        \end{eqnarray*}
iff
        $\lim_{\alpha \downarrow 1} \frac{x_{\alpha} - 1}{1 - \alpha}$
        exists. We next show the existence of this limit
        and verify that it is indeed equal to $h(p;q)$.
        Consider
        \begin{eqnarray*}
                \lim_{\alpha \downarrow 1} \frac{x_{\alpha} - 1}{1 - \alpha}
		&=& \lim_{\alpha \downarrow 1} \int_{\mathcal{R}} \frac{p\times q^{\alpha - 1} - p }{1 - \alpha}
                        \der \mu.
        \end{eqnarray*}
        In order to invoke the monotone convergence theorem,
        we prove that the integrand is non-decreasing and bounded below as $\alpha \downarrow 1$.
        Noting that
        \begin{eqnarray*}
                \frac{\der}{\der \alpha} \frac{p\times q^{\alpha - 1} - p}{1 - \alpha} 
		&=& \frac{p  q^{\alpha - 1} (1- (\alpha - 1) \log (q)) - p}{(\alpha - 1)^2},
        \end{eqnarray*}
        it is enough to show that
        \[p  q^{\alpha - 1} (1- (\alpha - 1) \log (q)) - p \leq 0.
        \]
        Indeed, we have
        \begin{equation}
		p [q^{\alpha - 1} (1 + \log (q^{1 - \alpha})) - 1]
                \leq p [q^{\alpha - 1} (1 + q^{1 - \alpha} - 1) - 1] 
                = p (1 - 1)
                = 0,
\label{eqn2}	\end{equation}
 where \eqref{eqn2} holds since $\log(x) \leq x - 1$, for $x>0$. 
        We next show that the integrand is bounded from below.  
	The lower bound can be obtained by letting $\alpha \rightarrow \infty$:
        \begin{eqnarray*}
                \lim_{\alpha \rightarrow \infty }\frac{p \times q^{\alpha - 1} - p}{1 - \alpha}
                &=& 0.
        \end{eqnarray*}
        Hence, by the monotone convergence theorem, we have that
        \begin{eqnarray*}
		\lim_{\alpha \downarrow 1} \int_{\mathcal{R}} \frac{p \times q^{\alpha - 1} - p}{1 - \alpha}
                        \der \mu 
		 = \int_{\mathcal{R}} \lim_{\alpha \downarrow 1} \frac{p \times q^{\alpha - 1} - p}{1 - \alpha}
                        \der \mu.
        \end{eqnarray*}
        Finally, by L'H\^{o}pital's rule, we obtain that
        \begin{equation*}
                \lim_{\alpha \downarrow 1} \frac{x_{\alpha} - 1}{1 - \alpha}
		= - \int_{\mathcal{R}} \frac{p \log(q)}{1}
                        \der \mu 
		= h(p;q).
        \end{equation*}

        Next, we shall prove taking $\alpha \uparrow 1$.
        For $\alpha < 1$,
        $
            x_{\alpha} = \E{\mathbf{A} \sim p}(q(\mathbf{A})^{\alpha - 1}) < \infty
        $
        by the fact that $ x_{\alpha}$
        is non-increasing in $\alpha > 0$ and by the assumption
        that $\E{\mathbf{A} \sim p}\left(\frac{1}{q(\mathbf{A})} \right) < \infty$.
        Hence, $\lim_{\alpha \uparrow 1} x_{\alpha} = 1$. Using Lemma~\ref{erwen}, we can
        apply the same steps as above with the alteration of the following argument.
        Consider
        \begin{eqnarray*}
                \lim_{\alpha \uparrow 1} \frac{x_{\alpha} - 1}{1 - \alpha}
                &=& \lim_{\alpha \uparrow 1} \int_{\X} \frac{p q^{\alpha - 1} - p}{1 - \alpha}
                        \der \mu.
        \end{eqnarray*}
        We know that the integrand is non-increasing in $\alpha > 0$.
        To use the monotone convergence theorem, we need to show
        that the integrand is bounded above. Note that
        \begin{equation*}
                \lim_{\alpha \rightarrow 0 }\frac{p \times q^{\alpha - 1} - p}{1 - \alpha}
                \ = \ \frac{p}{q} - p 
		\ \leq \ \frac{p}{q}.
        \end{equation*}
        Since we assumed $\E{\mathbf{A} \sim p}\left(\frac{1}{q(\mathbf{A})} \right) < \infty$, 
	we have that $\frac{p}{q} < \infty$ (a.e.).
        Thus the integrand is bounded above (a.e.).
        Following the same steps as in the previous part, we conclude that
        $\lim_{\alpha \uparrow 1} h_{\alpha}(p;q)=h(p;q)$.
        \qed

        We next show that the  \renyi \ cross-entropy is decreasing in $\alpha>0$, $\alpha \neq 1$.
\begin{theorem} 
        The \renyi \ cross-entropy $h_{\alpha}(p; q)$ between $p$ and $q$ is monotonically decreasing in $\alpha$, for $\alpha > 0$, $\alpha \neq 1$.
\end{theorem}

{\bf Proof }
First note that
\begin{eqnarray*}
	\frac{\der}{\der \alpha} h_{\alpha}(p; q) 
	&=& \frac{1}{(1 - \alpha)^2} \log \left(\int_{\mathcal{R}} p q^{\alpha - 1} \der \mu \right) \\
	&& \qquad	+ \frac{1}{1 - \alpha} \frac{1}{\int_{\mathcal{R}} p q^{\alpha - 1} \der \mu}
		\frac{\der}{\der \alpha} \left(\int_{\mathcal{R}} p q^{\alpha - 1} \der \mu \right). \nonumber
\end{eqnarray*}
We next consider
\begin{equation*}
	\frac{\der}{\der \alpha} \left(\int_{\mathcal{R}} p q^{\alpha - 1} \der \mu \right)
		= \lim_{\epsilon \downarrow 0} \int_{\mathcal{R}} p \left( \frac{q^{\alpha + \epsilon - 1} - q^{\alpha - 1}}{\epsilon} \right) \der \mu 
	= \lim_{\epsilon \downarrow 0} \int_{\mathcal{R}} p q^{\alpha - 1} \left( \frac{q^{\epsilon} - 1}{\epsilon} \right) \der \mu.
\end{equation*}
In order to apply the monotone convergence theorem, we prove the integrand is decreasing in $\epsilon \in (0, 1)$ as $\epsilon \downarrow 0$ and is bounded above.
Note that
\begin{eqnarray*}
	\frac{\der}{\der \epsilon} \frac{q^{\epsilon} - 1}{\epsilon} &=& \frac{1 + q^{\epsilon} (\epsilon \log(q) - 1)}{\epsilon^2},
\end{eqnarray*}
hence, it is enough to show that
	$1 + q^{\epsilon} (\epsilon \log(q) - 1) \geq 0.$
Indeed, we have
\begin{equation}
	1 + q^{\epsilon} (\log(q^{\epsilon}) - 1) \geq 1 + q^{\epsilon} \left(1 - \frac{1}{q^{\epsilon}} - 1 \right)  
	= 0, \label{eqn:mct_renyi_monotone}
\end{equation}
where~\eqref{eqn:mct_renyi_monotone} holds since $1 - \frac{1}{x} \leq \log(x)$, for all $x>0$.
We now show that the integrand is bounded above by letting $\epsilon \rightarrow 1$:
\begin{eqnarray*}
	\lim_{\epsilon \rightarrow 1} \frac{q^{\epsilon} - 1}{\epsilon} = q - 1 < \infty.
\end{eqnarray*}
Hence by the monotone convergence theorem we have
\begin{eqnarray*}
	\lim_{\epsilon \downarrow 0} \int_{\mathcal{R}} p q^{\alpha - 1} \left( \frac{q^{\epsilon} - 1}{\epsilon} \right) \der \mu
		= \int_{\mathcal{R}} \lim_{\epsilon \downarrow 0} p q^{\alpha - 1} \left( \frac{q^{\epsilon} - 1}{\epsilon} \right) \der \mu.
\end{eqnarray*}
By L'H\^{o}pital's rule, we obtain that
\begin{eqnarray*}
	\frac{\der}{\der \alpha} \left(\int_{\mathcal{R}} p q^{\alpha - 1} \der \mu \right) = \int_{\mathcal{R}} p q^{\alpha - 1} \log(q) \der \mu.
\end{eqnarray*}
Therefore, 
\begin{eqnarray*}
	\frac{\der}{\der \alpha} h_{\alpha}(p; q) &=&  \frac{1}{(1 - \alpha)^2} \log \left(\int_{\mathcal{R}} p q^{\alpha - 1} \der \mu \right) \nonumber\\
       && \qquad         + \frac{1}{1 - \alpha} \frac{1}{\int_{\mathcal{R}} p q^{\alpha - 1} \der \mu} \int_{\mathcal{R}} p q^{\alpha - 1} \log(q) \der \mu \\
	&=&  \frac{\int_{\mathcal{R}} p q^{\alpha - 1} (h_{\alpha}(p;q) + \log(q)) \der \mu}{(1 - \alpha)\left(\int_{\mathcal{R}} p q^{\alpha - 1} \der \mu \right)}.
\end{eqnarray*}
We next consider two cases, $\alpha < 1$ and $\alpha > 1$.
For $\alpha < 1$, we have
\begin{eqnarray}
	h_{\alpha}(p;q) + \log(q) &=& \frac{1}{1 - \alpha} \log\left(\int_{\mathcal{R}} p q^{\alpha - 1} \der \mu \right) + \frac{\log(q^{1 - \alpha})}{1 - \alpha} \nonumber \\
	&=& \frac{\log(q^{1 - \alpha} \int_{\mathcal{R}} p q^{\alpha - 1} \der \mu)}{1 - \alpha} \nonumber \\
	&\leq& \frac{q^{1 - \alpha} \int_{\mathcal{R}} p q^{\alpha - 1} \der \mu - 1}{1 - \alpha}, \label{eqn:log_lower_renyi_monotone}
\end{eqnarray}
where~\eqref{eqn:log_lower_renyi_monotone} holds since $\log(x) \leq x - 1$, for all $x > 0$.
We thus have
\begin{equation*}
	\frac{\der}{\der \alpha} h_{\alpha}(p; q)
		\leq \frac{\int_{\mathcal{R}} p q^{\alpha - 1}
		(q^{1 - \alpha} \int_{\mathcal{R}} p q^{\alpha - 1} \der \mu - 1) \der \mu}{(1 - \alpha)^2 \left(\int_{\mathcal{R}} p q^{\alpha - 1} \der \mu \right)} \\
	= \frac{\int_{\mathcal{R}} p q^{\alpha - 1} \der \mu - \int_{\mathcal{R}} p q^{\alpha - 1} \der \mu}
		{(1 - \alpha)^2 \left(\int_{\mathcal{R}} p q^{\alpha - 1} \der \mu \right)} 
	= 0.
\end{equation*}
Finally, for $\alpha > 1$, we consider
\begin{equation}
        h_{\alpha}(p;q) + \log(q) = \frac{\log(q^{1 - \alpha} \int_{\mathcal{R}} p q^{\alpha - 1} \der \mu)}{1 - \alpha} 
	\geq \frac{q^{1 - \alpha} \int_{\mathcal{R}} p q^{\alpha - 1} \der \mu - 1}{1 - \alpha}, \label{eqn:log_lower_renyi_monotone_2}
\end{equation}
where~\eqref{eqn:log_lower_renyi_monotone_2} holds since $-\log(x) \geq -(x - 1)$, for all $x > 0$.
Thus, we have
\begin{equation*}
        \frac{\der}{\der \alpha} h_{\alpha}(p; q)
                \leq \frac{\int_{\mathcal{R}} p q^{\alpha - 1}
                (q^{1 - \alpha} \int_{\mathcal{R}} p q^{\alpha - 1} \der \mu - 1) \der \mu}{(1 - \alpha)^2 \left(\int_{\mathcal{R}} p q^{\alpha - 1} \der \mu \right)} 
	= 0.
\end{equation*}
\qed

The above definition of differential
\renyi \ cross-entropy can be extended (assuming the integral exists)
by only requiring $q$ to be a non-negative function
(such as a non-normalized density function); in this case we call the resulting
measure as the (differential) {\em \renyi \ cross-entropy functional} and denote it by
$\mathcal{H}_{\alpha}(p;q)$. Similarly, we henceforth denote the Shannon cross-entropy functional by
$\mathcal{H}(p;q)$.

\begin{definition}\label{def: jensen-renyi-div}
The \textbf{Jensen-\renyi \ divergence of order} $\alpha$ between $p$ and $q$, 
where $\alpha > 0$, $\alpha \neq 1$, is given by
  {\rm       \begin{eqnarray} \label{jensenRenyiDivergenceFormula}
	\hspace{-0.17in}	\JR_{\alpha}(p \Vert q) &:=& \frac{1}{2} \D_{\alpha}\left( p \bigg\Vert \frac{p + q}{2} \right ) 
+ \frac{1}{2} \D_{\alpha}\left( q \bigg \Vert \frac{p + q}{2} \right).
	\end{eqnarray}
	}
\end{definition}

By the non-negativity of the \renyi \ divergence, it follows by definition
that $\JR_{\alpha}(p \Vert q) \ge 0$ with equality iff $p = q$ (a.e.).
Finally since $\lim_{\alpha \rightarrow 1} \D_{\alpha}(p \Vert q) = \KL(p \Vert q)$, we have that 
{\rm \begin{equation}
\lim_{\alpha \rightarrow 1} \JR_{\alpha}(p \Vert q) = \JSD(p \Vert q),
\end{equation}
}
where 
{\rm\begin{equation}\JSD(p \Vert q) := \frac{1}{2} \KL \left( p \bigg\Vert \frac{p + q}{2} \right ) 
+\frac{1}{2} \KL \left( q \bigg\Vert \frac{p + q}{2} \right )\end{equation}
}
is the Jensen-Shannon divergence.

%% file: lkgan.tex
\section{Least $\mathbf{k}$th-Order GANs (L$\mathbf{k}$GANs)}
\label{sec:lkgan}
\subsection{Theoretical results}
We generalize the LSGAN~\cite{lsgan} generator loss function based on the Pearson-Vajda divergence of order $k$, $\PV$, which subsumes the Pearson $\chi^2$ divergence. 
	Let	$(\X, \mathcal{B}(\X), \mu)$ be the measure space of $n \times n \times 3$ images, and
        $(\mathcal{Z}, \mathcal{B}(\mathcal{Z}), \mu)$ the measure space where $\mathcal{Z} \subset \real^{3n^2}$.
        Given discriminator neural network
        $D: \X \rightarrow [0, 1]$, and generator neural network  $g: \mathcal{Z} \to \X$, 
        let $p_{\mathbf{X}}: \X \to [0,1]$ be the 
        probability density function of real images and let $p_{\mathbf{Z}}: \mathcal{Z} \to [0,1]$ be the density function from which the generative neural network draws samples 
        (typically given by the multivariate Gaussian).
\begin{definition}
	The \textbf{Least $\mathbf{k}$th-order GANs loss functions}, $k \geq 1$, are defined as 
	\begin{eqnarray}
		V_D(D, g) &=& \frac{1}{2} \E{\mathbf{A} \sim p_{\mathbf{X}}} \left [ (D(\mathbf{A}) - b )^2 \right] 
                             + \frac{1}{2} \E{\mathbf{B} \sim p_{\mathbf{Z}}} \left [ ( D(g(\mathbf{B})) - a )^2 \right]  \\
                V_{k, g}(D, g) &=&  \E{\mathbf{A} \sim p_{\mathbf{X}}} \left ( \vert D(\mathbf{A}) - c  \vert^k \right)
                                + \E{\mathbf{B} \sim p_{\mathbf{Z}}} \left ( \vert D(g(\mathbf{B})) - c \vert^k \right), \label{lkgan-genf}
	\end{eqnarray}
	where $a$, $b$, $c \in [0, 1]$
	are constants,\footnote{More specifically, $a$ and $b$ are the discriminator's labels for fake and real data, respectively, while $c$ denotes the value that the generator aims the discriminator to believe for fake data.} and $V_D(D, g)$ and $V_{k, g}(D, g)$ are the discriminator and generator loss functions, respectively.
\end{definition}
Note that when $k = 2$ in \eqref{lkgan-genf}, the LSGANs generator's loss function is recovered. 
The resulting L$k$GAN problems are then to solve
\begin{eqnarray}
	 \min_D V_D(D, g) &=& \min_D \frac{1}{2} \E{\mathbf{A} \sim p_{\mathbf{X}}} \left [ (D(\mathbf{A}) - b )^2 \right] 
			     + \frac{1}{2} \E{\mathbf{B} \sim p_{\mathbf{Z}}} \left [ ( D(g(\mathbf{B})) - a )^2 \right] \label{eqn:dis_lkgan} \\
         \min_g V_{k, g}(D, g) &=&  \min_g \E{\mathbf{A} \sim p_{\mathbf{X}}} \left ( \vert D(\mathbf{A}) - c  \vert^k \right)
				+ \E{\mathbf{B} \sim p_{\mathbf{Z}}} \left ( \vert D(g(\mathbf{B})) - c \vert^k \right) \label{eqn:gen_lkgan}.
\end{eqnarray}

\begin{theorem} \label{theorem:lkgan}
        Consider problems~\eqref{eqn:dis_lkgan} and~\eqref{eqn:gen_lkgan} for training the discriminator and generator neural networks, respectively.
        Then
        \begin{eqnarray}
		D^* &:=& \argmin_D V_D(D, g) =  \frac{a p_g + b p_{\mathbf{X}}}{p_g + p_{\mathbf{X}}} \ \text{ (a.e.)} \label{lkgan1},
        \end{eqnarray}
	where $p_g: \X \to [0,1]$ is the generator's probability density. Furthermore, if $D = D^*$ and $a - b = 2 (c - b)$, 
	then 
	\begin{equation*}
		 V_{k, g}(D^*, g) =  \vert c - b \vert^{k} \PV(p_{\mathbf{X}} + p_g \Vert 2 p_g) \\
		 \geq 0,
	\end{equation*}
	with equality iff $p_g = p_{\mathbf{X}}$ (a.e.) or $c = b$.
\end{theorem}
\noindent
\textbf{Proof }
	The proof that the solution of \eqref{eqn:dis_lkgan} is 
	$D^* = \frac{a p_g + b p_{\mathbf{X}}}{p_g + p_{\mathbf{X}}}$ (a.e.) is presented in~\cite{lsgan}.
	Note that
		$\E{\mathbf{B} \sim p_{\mathbf{Z}}} 
		\left ( \vert D^*(g(\mathbf{B})) - c \vert^k \right) =  \E{\mathbf{B} \sim p_g} \left ( \vert D^*(\mathbf{B}) - c \vert^k \right).$	
	 Hence, 
	\begin{eqnarray*}
		V_{k, g}(D^*, g)  &=&  \E{\mathbf{A} \sim p_{\mathbf{X}}} \left ( \vert D^*(\mathbf{A}) - c \vert ^k \right) +
				\E{\mathbf{B} \sim p_g} \left ( \vert D^*(\mathbf{B}) - c \vert^k \right) \\
		&=&  \E{\mathbf{A} \sim p_{\mathbf{X}}} \left ( \left \vert 
				\frac{a p_g(\mathbf{A}) + b p_{\mathbf{X}}(\mathbf{A})}{p_g(\mathbf{A}) + p_{\mathbf{X}}(\mathbf{A})} - c 
				\right \vert ^k \right) + \E{\mathbf{B} \sim p_g} \left ( \left \vert 
				\frac{a p_g(\mathbf{B}) + b p_{\mathbf{X}}(\mathbf{B})}{p_g(\mathbf{B}) + p_{\mathbf{X}}(\mathbf{B})} - c 
				\right \vert ^k \right)  \\
		&=&  \int_{\X} p_{\mathbf{X}}\left \vert 
				\frac{(a - c) p_g+ (b - c) p_{\mathbf{X}}}{p_g+ p_{\mathbf{X}}} 
			\right \vert^k 
			\der \mu + \int_{\X} p_g \left \vert 
                                \frac{(a - c) p_g+ (b - c) p_{\mathbf{X}}}{p_g+ p_{\mathbf{X}}} 
                        \right \vert^k 
                        \der \mu  \\
		&=&  \int_{\X} 
			\frac{\left \vert (a - c) p_g + (b - c) p_{\mathbf{X}} \right \vert^k}
				{\left ( p_g + p_{\mathbf{X}} \right ) ^{k-1}}
			\der \mu \\
		&=&  \int_{\X}
                        \frac{\left \vert (a + b - b - c) p_g 
					+ (b - c) p_{\mathbf{X}} \right\vert^k}
                                {\left( p_g + p_{\mathbf{X}} \right)^{k-1}}
                        \der \mu \\
		&=&  \int_{\X}
                        \frac{\left \vert (a - b) p_g 
					+ (b - c) (p_{\mathbf{X}} + p_g) \right \vert^k}
                                {\left( p_g + p_{\mathbf{X}} \right)^{k-1}}
                        \der \mu. 
	\end{eqnarray*}
	Since $a - b = 2 (c - b)$, 
	we have that
	\begin{equation}
		V_{k, g}(D^*, g) =  \vert c - b \vert^{k}\int_{\X}
			\frac{ \left \vert 2 p_g
					- (p_{\mathbf{X}} + p_g) \right \vert^k}
				{\left( p_g + p_{\mathbf{X}} \right)^{k-1}}
                        \der \mu 
		=  \vert c - b \vert^{k} \PV(p_{\mathbf{X}} + p_g \Vert 2 p_g), \label{eqn:lkgan_results}
	\end{equation}
	which is minimized when $p_{\mathbf{X}} + p_{g} = 2 p_{g}$, equivalently $p_{\mathbf{X}} = p_{g}$ (a.e.), or 
	when $c = b$.
	\qed \\

L$k$GANs confer an extra degree of freedom by virtue of the order $k \geq 1$. 
Moreover, the minimum is theoretically achieved when 
the generator's distribution is the true distribution, which is same as LSGANs.

\subsection{Experiments}
\subsubsection{Methods} \label{sec: lkgan-methods}
The CelebA \cite{celeba} dataset was used to test the new L$k$GANs loss functions.
For comparison, LSGANs were also implemented.
The publicly available StyleGAN code from \cite{styleGAN} was modified to incorporate the $k$th-order generator loss function; the resulting system is referred to as L$k$StyleGAN. The architectures of the generator and discriminator neural networks are described in  Section~\ref{architectures} and Algorithm~\ref{algo:lkgan} in the Appendix for details.\footnote{\label{mnist-footnote}We did also implement the L$k$GAN loss functions on the MNIST database under the DCGAN architecture. The results, which we do not report here, followed in general a similar trend to the L$k$StyleGAN CelebA results. }
The FID score was used to evaluate the quality of the generated images and to compare the rate at which the new networks converged to their optimal FID scores.
Due to the large computational demand of the StyleGAN architecture, we only performed three trials with seeds 1000, 2000, and 3000 for trials 1, 2, and 3, respectively. 
L$k$StyleGANs were implemented for $k \in \{1, 2, 3\}$.
We refer to L$k$StyleGANs when $k=2$ as LSStyleGANs.
Three variants of L$k$StyleGANs with differing $a$, $b$, and $c$ parameters were tested. 
Version $1$, L$k$StyleGAN-v1, has $a = 0.6$, $b = 0.4$, and $c = 0.5$.
Version $2$, L$k$StyleGAN-v2, has $a = 1$, $b = 0$, and $c = 0.5$.
Version $3$, L$k$StyleGAN-v3, has $a = 0$, $b = 1$, and $c = 1$, which are the parameters tested in \cite{lsgan}. 
L$k$StyleGANs with and without simplified gradient penalties were also implemented.
We denote L$k$StyleGANs with simplified gradient penalties as L$k$StyleGAN-GP.

The original StyleGAN architectural defaults were left in place for L$k$StyleGAN. 
The batch size was chosen to be~128. 
Unlike the original StyleGAN, which changes the resolution of its generated images during training,
the resolution of the generated images for L$k$StyleGAN was fixed at $64 \times 64 \times 3$ throughout training. 
As recommended by the original StyleGANs paper, 
the Adam optimizer with a learning rate of $\alpha_{Adam}=0.001$, $\beta_1=0.0, \beta_2=0.99$ and $\epsilon=10^{-8}$ 
was used as the stochastic gradient descent (SGD) algorithm \cite{styleGAN}. 
The L$k$StyleGAN systems were trained for 25 million images or roughly 120 epochs using
one NVIDIA GP$100$ GPU and two Intel Xeon $2.6$ GHz E$7-8867$ v$3$ CPUs.
We will say that a network has \textit{training stability} if it converges to meaningful results
(i.e., it does not suffer from mode collapse).


\subsubsection{Results}
L$k$StyleGANs with and without the simplified gradient penalty were tested over three trials while controlling the 
seed in each trial. 
The addition of the simplified gradient penalty significantly increased training time. 
Hence only Version~2 of L$k$StyleGAN-GP was tested for $k=\{1, 2, 3\}$.
The FID scores were calculated every $80,000$ images. 
The average and variance of best FID scores taken over the three trials are presented in Table \ref{table:FID average lkstylegan}.
We also plot the average FID scores taken over the three trials 
versus epochs in Figure~\ref{fig:lkstylegan average plots}.
Finally, sample generated images of one of the trials for L$k$StyleGANs and LSStyleGANs are shown in Figure~\ref{fig:lkstylegan generated images trial 2}.

\subsection{Discussion}
The choice of $k=1$ produced the best performing L$k$StyleGAN. 
In contrast, LSStyleGANs and L$k$StyleGANs-3.0 suffered from mode collapse in all three trials for both Versions~1 and~2. 
For Version~3, LSStyleGAN converged to meaningful results, however, as training continued, it suffered from mode collapse.

Similarly, L$k$StyleGAN-v1-1.0 and L$k$StyleGAN-v3-1.0 converged to meaningful results in most trials and suffered from mode 
collapse as training continued. 
The exception to this is trial~2 of L$k$StyleGAN-v3-1.0, which exhibited training stability. 
Note that, L$k$StyleGAN-v2-1.0 only converged in the two out of three trials. 
This supports the fact that the choice of $a$, $b$, and $c$ parameters plays a pertinent role
in the networks' convergence behaviours. 
Note that L$k$StyleGAN-v1-1.0 generated image quality was superior to 
that of the generated images of convergent L$k$StyleGAN-v2-1.0.
Furthermore, L$k$StyleGAN-v3-1.0 and LSStyleGAN-v3 converged to meaningful results and outperformed the other versions.
This implies that the conditions provided by
Theorem~\ref{theorem:lkgan} do not need to be satisfied for certain network architectures.

\begin{figure}[htb]
    \centering
        \begin{subfigure}[t]{.47\textwidth}
         \centering
        \includegraphics[width=75mm]{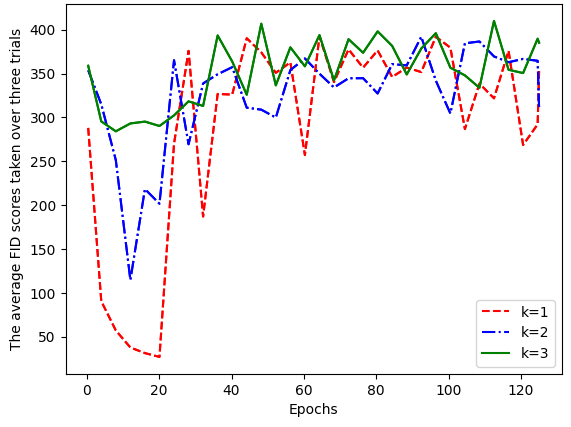}
        \caption{Average FID vs epochs, L$k$StyleGAN-v1.}
        \label{fig:lkganv1}
     \end{subfigure}
	\quad
	\begin{subfigure}[t]{0.47\textwidth}
         \centering
             \includegraphics[width=75mm]{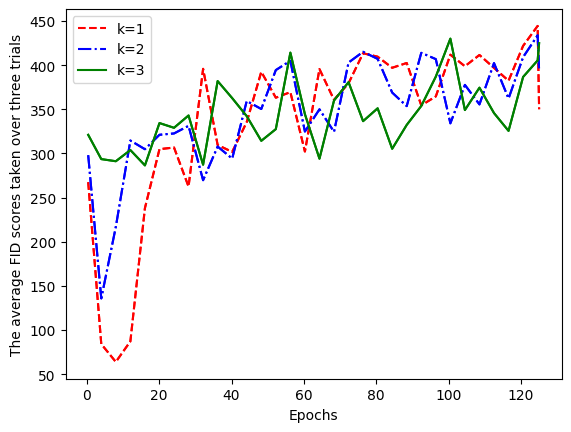}
        \caption{Average FID vs epochs, L$k$StyleGAN-v2.}
        \label{fig:lkganv2ave}
     \end{subfigure}
     \quad
     	\begin{subfigure}[t]{0.47\textwidth}
         \centering
             \includegraphics[width=75mm]{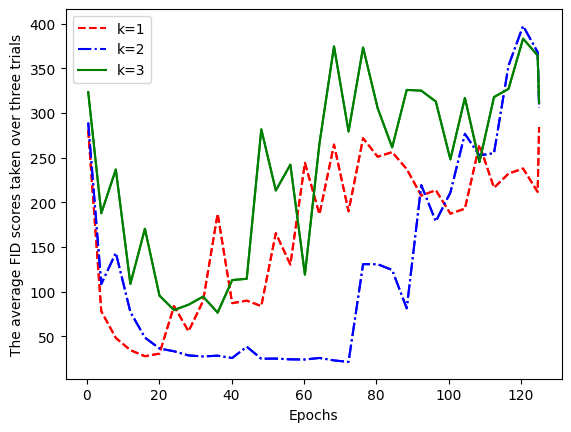}
        \caption{Average FID vs epochs, L$k$StyleGAN-v3.}
        \label{fig:lkganv3ave}
     \end{subfigure}
     \quad
	\begin{subfigure}[t]{0.47\textwidth}
         \centering
	     \includegraphics[width=75mm]{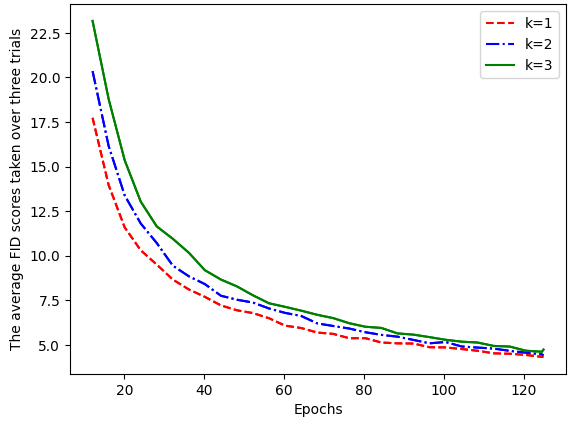}
        \caption{Average FID vs epochs, L$k$StyleGAN-v2-GP.}
        \label{fig:lkganv2gpave}
     \end{subfigure}
     \quad
    \caption{Evolution of the average FID scores throughout training for L$k$StyleGANs on CelebA (64 $\times$ 64) dataset.}
    \label{fig:lkstylegan average plots}
\end{figure}

Without the simplified gradient penalty, these networks took~87.87 hours to train for one trial on one GPU. 
The addition of the simplified gradient penalty increased training time to~106.35 hours, an increase of 21\%.
However, the simplified gradient penalty improved the quality of generated images throughout training for all three trials, as seen in Figure \ref{fig:lkganv2gpave}.
This confirms our hypothesis that the discriminator creates gradients 
despite the fact that the generated images are close to real images.
The average FID scores increased with $k$, as with the previous experiments; see Figure~\ref{fig:lkganv2gpave}.

In summary, the choice of the $a$, $b$, and $c$ parameters has a significant effect on training stability and quality of generated images,
and the best choice of these parameters differs for different network architectures.
However, when L$k$GANs,~$k = 1$, do converge, they consistently improved the quality of 
generated images in terms of FID scores, 
converged to their optimal FID score quicker, and  
improved training stability compared to their counterpart, LSGANs.
Furthermore, they give rise to interesting theoretical problems and experiments for future research.


\begin{table}[hb]
\caption{L$k$StyleGANs experiments on the (64 $\times$ 64) CelebA dataset: the average and variance of the best FID score and the average and variance 
	epoch this occurs taken over three trials.}
        \label{table:FID average lkstylegan}
\vskip 0.1in
\centering
\begin{tabular}{L C N C N}
\hline \hline
	& Average best FID score & Best FID score variance & Average epoch & Epoch variance \\
 \hline
	{\bf L$\bm{k}$StyleGAN-v1-1.0} & {\bf 26.94} & {\bf 3.43} & 20.10 & {\bf 0.00} \\
	L$k$StyleGAN-v1-3.0 & 224.49 & 877.99 & 26.79 & 154.36 \\
	LSStyleGAN-v1 & 107.98 & 138.98 & {\bf 14.73} & 14.36 \\
        \hline
	{\bf L$\bm{k}$StyleGAN-v2-1.0} & {\bf 58.72} & 716.46 & {\bf 10.71} & {\bf 3.59} \\
	L$k$StyleGAN-v2-3.0 & 238.83 & {\bf 405.79} & 56.27 & 904.61 \\
	LSStyleGAN-v2 & 117.80 & 444.99 & 14.73 & 154.36 \\
	\hline
	{\bf L$\bm{k}$StyleGAN-v3-1.0} & {\bf 18.83 } &  {\bf 37.75} &  56.26 &  2358.60 \\
	L$k$StyleGAN-v3-3.0 &  62.80 &  415.01 & {\bf 28.13} & 172.32 \\
	LSStyleGAN-v3 & 20.46 & 82.89 & 79.04 & {\bf 25.12} \\
		\hline
	{\bf L$\bm{k}$StyleGAN-v2-1.0-GP} & \bf{4.31} & { $\mathbf{5.60 \times 10^{-3}}$} &  124.73 & { $\mathbf{3.64 \times 10^{-2}}$} \\
	L$k$StyleGAN-v2-3.0-GP & 4.59 & 3.54 $\times 10^{-2}$ & {\bf 123.26} & 3.59 \\
	LSStyleGAN-v2-GP & 4.42 & 6.34 $\times 10^{-2}$ & 124.87 & {$\mathbf{3.64 \times 10^{-2}}$} \\
	 \hline
\end{tabular}
\end{table}

\begin{figure}[hb] 
	\centering
        \begin{subfigure}[t]{0.47\textwidth}
                \centering
                \includegraphics[width=70mm]{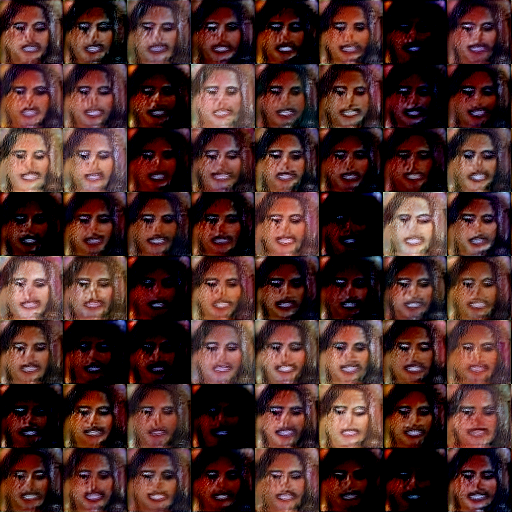}
                \caption{LSStyleGAN-v1: FID score~121.69.}\label{fig:lsgan_v1_t2_images}
        \end{subfigure}
        \quad
        \begin{subfigure}[t]{0.47\textwidth}
                \centering
                \includegraphics[width=70mm]{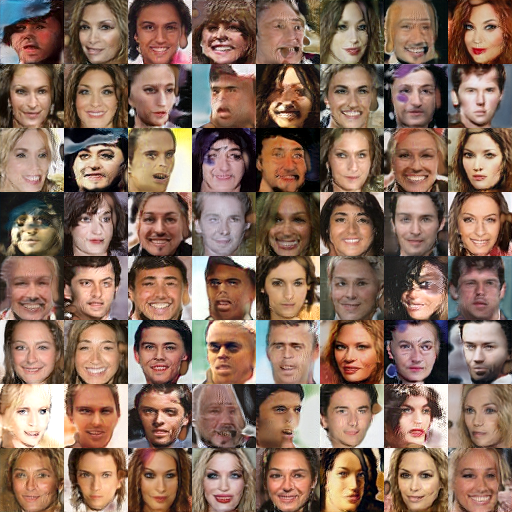}
                \caption{L$k$StyleGAN-v1-1.0: FID score~24.89.}\label{fig:lkgan_v1_t2_images}
        \end{subfigure}
        \quad
	\begin{subfigure}[t]{0.47\textwidth}
                \centering
                \includegraphics[width=70mm]{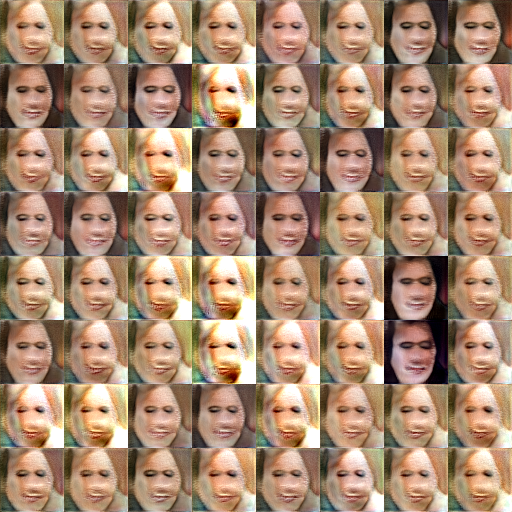}
                \caption{LSStyleGAN-v2: FID score~136.18.}\label{fig:lsgan_v2_t2_images}
        \end{subfigure}
        \quad
        \begin{subfigure}[t]{0.47\textwidth}
                \centering
                \includegraphics[width=70mm]{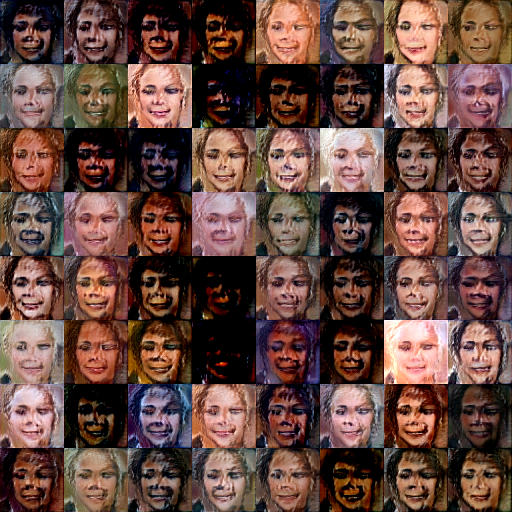}
                \caption{L$k$StyleGAN-v2-1.0: FID score~96.56.}\label{fig:lkgan_v2_t2_images}
        \end{subfigure}
        \quad
\end{figure}
\begin{figure}[htb] \ContinuedFloat
	\centering
	\begin{subfigure}[t]{0.47\textwidth}
                \centering
                \includegraphics[width=70mm]{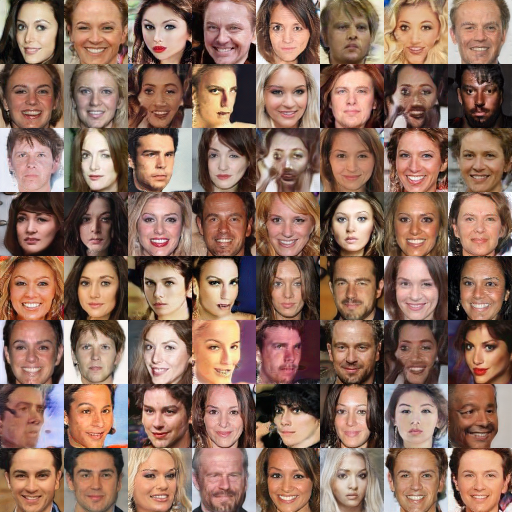}
                \caption{LSStyleGAN-v3: FID score~17.48.}\label{fig:lsgan_v3_t2_images}
        \end{subfigure}
        \quad
        \begin{subfigure}[t]{0.47\textwidth}
                \centering
                \includegraphics[width=70mm]{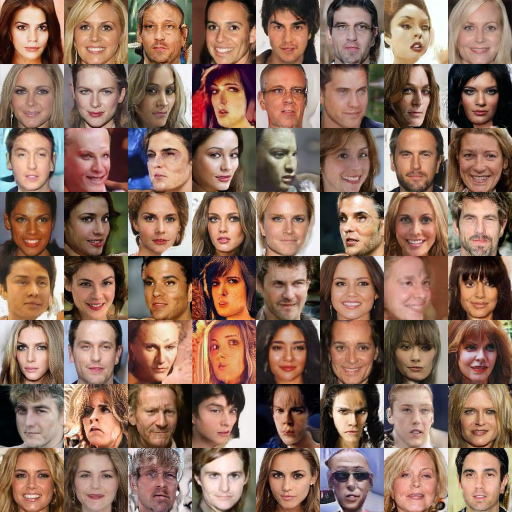}
                \caption{L$k$StyleGAN-v3-1.0: FID score~12.32.}\label{fig:lkgan_v3_t2_images}
        \end{subfigure}
        \quad
        \begin{subfigure}[t]{0.47\textwidth}
                \centering
                \includegraphics[width=70mm]{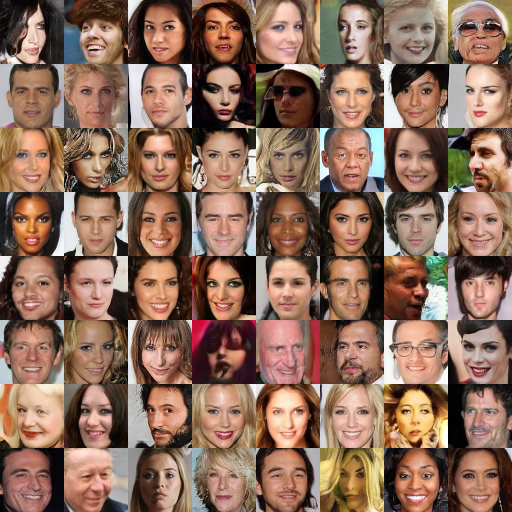}
                \caption{LSStyleGAN-v2-GP: FID score~4.52.}\label{fig:lsgan_v2gp_t2_images}
        \end{subfigure}
        \quad
        \begin{subfigure}[t]{0.47\textwidth}
                \centering
                \includegraphics[width=70mm]{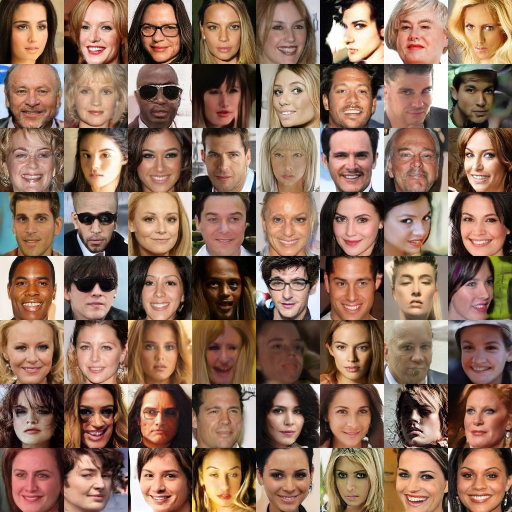}
		\caption{L$k$StyleGAN-v2-1.0-GP:  FID score~4.21.}\label{fig:lkgan_v2gp_t2_images}
        \end{subfigure}
        \quad
        \caption{CelebA (64~$\times$~64)  generated images of the best performing L$k$StyleGANs
        for each version and their LSStyleGAN counterparts for a sample trial. }
        \label{fig:lkstylegan generated images trial 2}
\end{figure}

\clearpage

%% file: renyigan.tex
\section{\renyi GANs}
\label{sec:renyigan}
\subsection{Theoretical results}
Motivated by generalizing the original GANs, 
we employ the \renyi \ cross-entropy loss functional and Jensen-\renyi \ divergence 
of order $\alpha > 0$, $\alpha \neq 1$, which generalizes
the Shannon cross-entropy functional and Jensen-Shannon divergence.  
Using the same setup and notations as in Section~\ref{sec:lkgan}, we next
present the \renyi GANs loss functions. 
\begin{definition}
	The \textbf{\renyi GANs loss functions of order $\bm{\alpha}$}, $\alpha > 0$, $\alpha \neq 1$, 
	are defined as 
\begin{eqnarray}
	V_D(D,g) &=& -\mathcal{H}(p_{\mathbf{X}};D) - \mathcal{H}(p_{\mathbf{Z}}; 1 - D \circ g) \label{renyigan1} \\
	 V_{\alpha, g}(D, g) &=& - \mathcal{H}_{\alpha}(p_{\mathbf{X}}; D) - \mathcal{H}_{\alpha}(p_{\mathbf{Z}}; 1 - D\circ g),  \label{renyigan}
\end{eqnarray}
	where $\circ$ denotes functional composition, and $\mathcal{H}(\cdot)$ and $\mathcal{H}_{\alpha}(\cdot)$ denote the Shannon and \renyi \ cross-entropy functionals, respectively (see end of Section~\ref{sec:divMeasures} before Definition~\ref{def: jensen-renyi-div}).
\end{definition}
Note that for \renyi GANs, $V_D(D,g) = V(D, g)$, where $V(D, g)$ is the original
GANs loss function as defined in~\cite{Goodfellow2014}. 
The resulting \renyi GANs optimization problems consist then of determining:
\begin{eqnarray}
	\max_D V_D(D,g) &=& \max_D \left(-\mathcal{H}(p_{\mathbf{X}};D) - \mathcal{H}(p_{\mathbf{Z}}; 1 - D \circ g) \right) \label{renyigan_dis_opt}\\
	\min_g V_{\alpha, g}(D, g) &=&  \min_g \left (-\mathcal{H}_{\alpha}(p_{\mathbf{X}}; D) - \mathcal{H}_{\alpha}(p_{\mathbf{Z}}; 1 - D\circ g) \right), \label{renyigan_gen_opt}
\end{eqnarray}
The \renyi GANs' generator tries to induce the discriminator to classify the fake images 
as $1$ by minimizing 
the negative sum of the two {\em \renyi \ cross-entropy functionals} 
$\mathcal{H}_{\alpha}(p_{\mathbf{X}}; D)$ and $\mathcal{H}_{\alpha} (p_{\mathbf{Z}}; 1 - D\circ g)$,
hence generalizing the original GANs loss function by employing a richer $\alpha$-parameterized class of information functionals.
Indeed we have that as $\alpha \rightarrow 1$, we recover the original GANs loss function, $V(D, g)$; 
this is formalized in the following result, whose proof follows directly from Theorem~\ref{theorem:renyi-crossentropy-limit}. 
\begin{theorem}\label{theorem:renyi-limit}
	If $V(D, g) < \infty$, then 
	\begin{eqnarray}
		\lim_{\alpha \downarrow 1} V_{\alpha, g}(D,g) &=& V(D,g).
	\end{eqnarray}
	Moreover, 
	if $\E{\mathbf{A} \sim p_{\mathbf{X}}} \left( \frac{1}{D(\mathbf{A})} \right ) < \infty$ 
	and $\E{\mathbf{B} \sim p_{\mathbf{Z}}} \left( \frac{1}{1 - D(g(\mathbf{B}))} \right ) < \infty$
	then
	\begin{eqnarray}
		\lim_{\alpha \uparrow 1} V_{\alpha, g}(D,g) &=& V(D,g).\label{eq:lower}
	\end{eqnarray}	
\end{theorem}

If the discriminator converges to the optimal discriminator, we next show analytically that for any $\alpha > 0$, $\alpha \neq 1$,
the optimal generator induces a probability distribution that perfectly mimics the true 
dataset distribution, as in GANs. 
\begin{theorem} \label{theorem:min_max_pg}
	Consider problems~\eqref{renyigan_dis_opt} and~\eqref{renyigan_gen_opt} for training the discriminator and generator neural networks, respectively.
        Then
        \begin{eqnarray}
                D^* &:=& \argmax_D V_D(D, g) = \frac{p_{\mathbf{X}}}{p_g + p_{\mathbf{X}}} \ \text{ (a.e.)}.
        \end{eqnarray}
        Furthermore, if $D = D^*$, then
	\begin{equation*}
		V_{\alpha, g}(D^*, g) = 2\JR_{\alpha} \left( p_{\mathbf{X}} \Vert p_g \right) -2 \log(2) 
		\geq -2\log(2),
	\end{equation*}
	with equality iff $p_g = p_{\mathbf{X}}$ (a.e.). 
\end{theorem}

\noindent
{\bf Proof }
        The proof that the solution to \eqref{renyigan_dis_opt} is
        $D^* = p_{\mathbf{X}}/(p_{\mathbf{X}}+p_g)$ is in~\cite{Goodfellow2014}.
	We have that
	$	\E{\mathbf{B} \sim p_{\mathbf{Z}}} \left[ \left(1 - D^*(g(\mathbf{B})) \right)
                        ^{\alpha - 1}\right] 
		= \E{\mathbf{B} \sim p_{\mathbf{g}}} \left[ \left(1 - D^*(\mathbf{B})) \right)
                        ^{\alpha - 1}\right].$ 
	Hence,
	\begin{eqnarray*}
		V_{\alpha, g}(D^*, g)&=& \frac{1}{\alpha - 1} \left[\log \left (
                        \E{\mathbf{A} \sim p_{\mathbf{X}}}\left[\left(D^*(\mathbf{A}) \right)^{\alpha - 1}\right] \right)
                                + 
                                \log \left (
                        \E{\mathbf{B} \sim p_g} \left[ \left(1 - D^*(\mathbf{B}) \right)
                        ^{\alpha - 1}\right]
                        \right)\right] \\
                &=&  \frac{1}{\alpha - 1} \log \left (
                        \E{\mathbf{A} \sim p_{\mathbf{X}}}\left[\left(\frac{2 p_{\mathbf{X}}(\mathbf{A})}{p_{\mathbf{X}}(\mathbf{A}) 
			+ p_g(\mathbf{A})}\right)^{\alpha - 1}\right] \right)  \\
                        && + \frac{1}{\alpha - 1} \log \left ( \E{\mathbf{B} \sim p_g} 
			\left[ \left(\frac{2 p_g(\mathbf{B})}{p_{\mathbf{X}}(\mathbf{B}) + p_g(\mathbf{B})}\right)
                        ^{\alpha - 1}\right]
                        \right) - 2 \log(2) \\
                &=&  2 \left[ \frac{1}{2} \D_{\alpha} \left( p_{\mathbf{X}} \bigg \Vert \frac{p_{\mathbf{X}} + p_g}{2} \right) +
                        \frac{1}{2} \D_{\alpha} \left( p_g \bigg \Vert \frac{p_{\mathbf{X}} + p_g}{2} \right) \right]
                                - 2 \log(2) \\
                &=&  2\JR_{\alpha} \left( p_{\mathbf{X}} \Vert p_g \right) -2 \log(2), 
        \end{eqnarray*}
	which is minimized when $p_{g} = p_{\mathbf{X}}$ (a.e.).
        \qed

\smallskip
This theorem implies that the introduction of the new loss function does not alter 
the underlying global equilibrium point of 
\renyi GANs when compared to the classical GANs 
(which use a Shannon-centric loss function), 
namely that the minimum is theoretically achieved when the generator's distribution is the true dataset distribution.
Using the above \renyi-centric loss function allows control of the shape of the 
generator's loss function via the order $\alpha$. 

\vspace{0.04in}
\noindent
{\em Loss function stability:}
We close this section by determining the {\em absolute condition number}~\cite{rgan} for our \renyi \ cross-entropy functional $\mathcal{H}_{\alpha}(\cdot;\cdot)$ used in the \renyi GAN generator loss function in \eqref{renyigan}.
Such quantity may be a useful to measure loss function stability; in this sense, the loss function is called ``stable'' if small changes in its argument lead to small changes in its absolute condition number. As our loss function in \eqref{renyigan} consists of the (negative) sum of two \renyi \ cross-entropy functionals, we hence focus on deriving the absolute condition number of our cross-entropy functional $\mathcal{H}_{\alpha}(p;\cdot)$ where $p$ is a fixed probability density with support $\mathcal{R}$. More precisely, the absolute condition number, $\kappa(F)(f(x_0))$, 
of a (Fr\'{e}chet differentiable) functional $F[f]$ at $f(x_0)$ (for $x_0 \in \mathcal{R}$) is defined as~\cite{rgan}
$$
\kappa(F)(f(x_0)) = \left| \frac{\delta F}{\delta f(x_0)}\right|
$$
where $\delta F/\delta f(x)$ is the functional derivative 
defined via 
\begin{equation*}
    \int \frac{\delta F[f]}{\delta f(x)} \eta(x) \der x
    :=  \frac{\der }{\der \epsilon} F[f + \epsilon \eta] \Big \vert_{\epsilon = 0 }
\end{equation*}
where $\epsilon \in \real$ and $\eta$ is an arbitrary function (see~\cite[Equation~(A.15)]{density_functional_theory}). 
Using similar calculations as in~\cite{rgan}, 
we obtain that the absolute condition number of our \renyi \ cross-entropy $\mathcal{H}_{\alpha}(p;q(x_0)) =:F_\alpha(q(x_0))$ at $q(x_0)$ is given by
\begin{equation}\label{cond-numb}
\kappa(F_\alpha)(q(x_0))) = \frac{p(x_0)q^{\alpha - 2}(x_0)}{\int_{\mathcal{R}} p(x) q^{\alpha - 1}(x) \der \mu},
\end{equation}
for $\alpha > 0$, $\alpha \neq 1$. 
For $\alpha < 2$, the absolute condition number in \eqref{cond-numb} approaches infinity as $q(x_0) \rightarrow 0$.
This implies that the generator's loss function for \renyi GAN in~\eqref{renyigan} produces large gradient updates if $D(x)$ or $1-D(g(z))$ are small. 
However, for $\alpha \geq 2$, the absolute condition number is bounded and hence the generator's gradient updates are stable for any $D(x)$ and $1 - D(g(z))$. 
In contrast, it is shown in~\cite{rgan} that the Shannon cross-entropy (which is used in the original GAN loss function, see~\eqref{renyigan1}) is unstable if $D(x)$ or $1-D(g(z))$ are small. Finally as noted in~\cite{rgan}, we emphasize that while we have ascertained a notion of stability for our  \renyi-type loss function for $\alpha \ge 2$ (which was observed in our experiments), there are other factors that play an important role in a GAN's overall stability (including the system architecture and the employed gradient decent technique). 

\subsection{Experiments}
\subsubsection{Methods}
The $28 \times 28 \times 1$ MNIST~\cite{mnist}, $64 \times 64 \times 3$ CelebA, and $128 \times 128 \times 3$ CelebA~\cite{celeba} datasets were used to test the \renyi GANs loss functions.
As in Section~\ref{sec:lkgan}, FID scores were used to evaluate the quality of the generated images and to compare the rate at which the new networks converge to their optimal scores.
The structure of the generator and discriminator neural networks were kept constant when testing on each dataset; see Section~\ref{architectures} and Algorithm~\ref{algo:renyigan} in the Appendix for details. 
For comparison, the classical (Shannon-centric) GANs loss functions were also tested. 

We denote \renyi GAN-$\alpha$ as \renyi GANs that use a fixed value value of $\alpha$ during training (see Algorithm \ref{algo:renyigan} in the Appendix). 
For the MNIST dataset, in addition to implementing \renyi GAN-$\alpha$, \renyi GANs were tested while altering $\alpha$ for every epoch of the simulation.
This changes the shape of the loss function of the generator.
However, changing $\alpha$ does not affect the global minimum as for all $\alpha > 0$, the global minimum is realized when $p_{\mathbf{X}} = p_g$.
Assuming that a generator $p_g \neq p_{\mathbf{X}}$ is realized such that it is not a local minimum of $V_{\alpha}(D, g)$ for all $\alpha >0$, then changing $\alpha$ every epoch creates non-zero gradients at previous local minimums, hence helping the algorithm overcome the problem of getting stuck in local minimums.
We denote this as \renyi GAN-$[\beta_1, \beta_2]$, with the $\alpha$ value starting at $\alpha = \beta_1$ and ranging over the interval $[\beta_1 ,\beta_2]$. 

One goal was to examine whether the generalized (\renyi-based) loss functions have
appreciable benefits over the classical GANs loss function and whether it provides better training stability.
It is known that deep convolutional GANs (DCGANs) exhibit stability issues which 
motivate us to investigate modifications to the loss functions involving the addition of the $L_1$ norm. 
Specifically, those stability issues arise when the GANs generator tries to minimize its cost function to $- \infty$ by labelling $D(g(z)) = 1$ for all fake images $g(z)$.
In the early stages of the simulations, if the discriminator does not successfully converge to its optimal value 
and the generator is able to induce the discriminator to label poorly generated images as $1$, 
then in later epochs, once the discriminator converges to its optimal value and 
is able to tell apart real and fake images perfectly, the generator's loss function produces no gradients.
In other words, the optimal discriminator does not allow the generator to improve the quality of fake images which leads to the discriminator winning problem.
A similar argument was noted in~\cite{wasserstein2017}.
Thus to remedy the stability problem, a modified \renyi GANs' generator's loss function was tested 
by taking the $L_1$ norm of its deviation from $-2\log (2)$, its theoretically
minimal value predicted by Theorem~\ref{theorem:min_max_pg}; this yields the following minimization problem for the generator network:
\begin{eqnarray}
\hspace{-0.22in}
\min_g \hspace{-0.02in} \Big| V_{\alpha, g}(D,g) -(-2\log 2) \Big| \label{l1-norm}
\end{eqnarray}
Using the $L_1$ norm ensures that the generator's loss function does not try to label its images as $1$, but rather tries to label them as $1/2$.
Hence in the early training stages, if the generator converges to images that are labelled $1/2$ by the discriminator, 
then in the later stages, if the discriminator converges to its theoretical optimal value 
(given in Theorem~\ref{theorem:min_max_pg}), the generator's loss function has 
non-zero gradient updates and is only able to label fake images as $1/2$ when $p_g = p_{\mathbf{X}}$.
Note that the altered loss function in~\eqref{l1-norm}  
translates into composing the L$k$GAN error function using $k=1$ and $\gamma=-\log(2)$ 
(see Section~\ref{sec:lkgan}) with the R\'enyiGAN loss function.  
Indeed, the improved stability property of L$k$GANs 
(particularly when $k=1$) is the main motivation for using this $L_1$ normalization.
We denote the resulting scheme under \eqref{l1-norm} by \renyi GAN-$L_1$.
The \renyi GANs and classical GANs loss functions were also tested with and without the addition of simplified gradient penalties. 
We denote this as \renyi GAN-GP and DCGAN-GP. 

In summary, we considered the evaluation of four versions of the algorithm with six different loss functions within each version.
Version $1$ has \renyi GAN-$0.5$, \renyi GAN-$3.0$, \renyi GAN-$[0, 0.9]$, \renyi GAN-$[0, 3.0]$, \renyi GAN-$[1.1, 4]$, and DCGAN.
Version $2$ has the six original loss functions with $L_1$ normalization,
Version $3$ has gradient penalty, and 
Version $4$ has gradient penalty and $L_1$ normalization incorporated in the loss functions.

For the MNIST dataset, seeds $123$, $5005$, $1600$, $199621$, $60677$, $20435$, $15859$, $33764$, $79878$, $36123$
were used for trials $1$ to $10$, respectively.
For the SGD algorithm,
the Adam optimizer with a learning rate of $\alpha_{Adam} = 2 \times 10^{-4}$, $\beta_1 = 0.5$, $\beta_2 = 0.999$, and
$\epsilon = 1 \times 10^{-7}$ was used for the networks as recommended by \cite{Radford2015}.
The batch size was chosen to be $100$ for the $60,000$ MNIST images.
The networks were trained on the MNIST dataset for a total number of $250$ epochs, or 15 million images.

For CelebA, as in Section~\ref{sec:lkgan}, we ran three trials with seeds 1000, 2000, and 3000 for trials 1, 2, and 3 respectively.
For comparison, the original StyleGAN with the classical GANs loss function was implemented. 
The publicly available StyleGAN code from \cite{styleGAN} was modified to test the \renyi GANs loss functions.
We refer to this as \renyi StyleGANs.
\renyi StyleGANs were implemented for $\alpha \in \{3.0, 9.0\}$.
\renyi StyleGANs and StyleGANs with the simplified gradient penalty, referred to as \renyi StyleGAN-GP and StyleGAN-GP, respectively,  were also tested.
The original StyleGAN architectural defaults were left in place with a batch size of~128.
As in Section~\ref{sec: lkgan-methods}, 
the resolution of the generated images for \renyi StyleGANs and StyleGAN was fixed at $64 \times 64 \times 3$ or at $128 \times 128 \times 3$ throughout training.
The same Adam optimizer parameters were used as specified in Section~\ref{sec: lkgan-methods} and the systems  were trained for 25 million images (roughly 120 epochs).
One NVIDIA GP$100$ GPU and two Intel Xeon $2.6$ GHz E$7-8867$ v$3$ CPUs were used for \renyi GANs and DCGANs,
and four NVIDIA V~100 GPUs were used for \renyi StyleGANs and StyleGANs.


\subsubsection{MNIST results}
\label{mnist results}
\begin{figure}[htb]
    \begin{subfigure}[t]{0.47\textwidth}
        \centering
        \includegraphics[width=70mm]{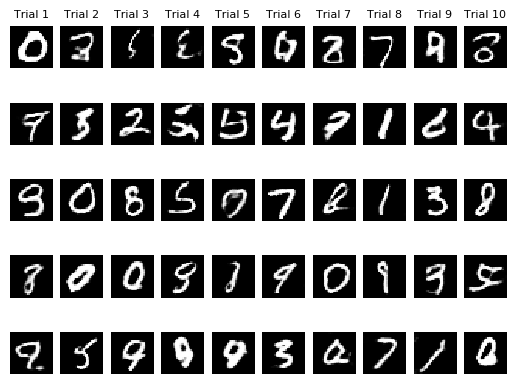}
        \caption{DCGAN-$L_1$ sample images.}\label{fig:dcgan_l1}
    \end{subfigure}
    \quad
    \begin{subfigure}[t]{0.47\textwidth}
        \centering
        \includegraphics[width=70mm]{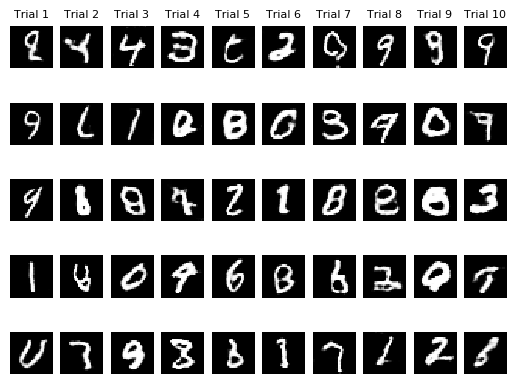}
        \caption{\renyi GAN-$[0, 3]$-$L_1$ sample images.}\label{fig:renyigan_2}
    \end{subfigure}
    \quad
	\centering
	\begin{subfigure}[t]{0.47\textwidth}
        \centering
        \includegraphics[width=70mm]{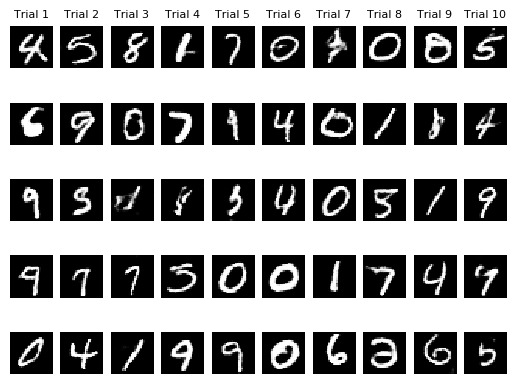}
        \caption{DCGAN-GP sample images.}\label{fig:dcgan_gp}
    \end{subfigure}
    \quad
    \begin{subfigure}[t]{0.47\textwidth}
        \centering
        \includegraphics[width=70mm]{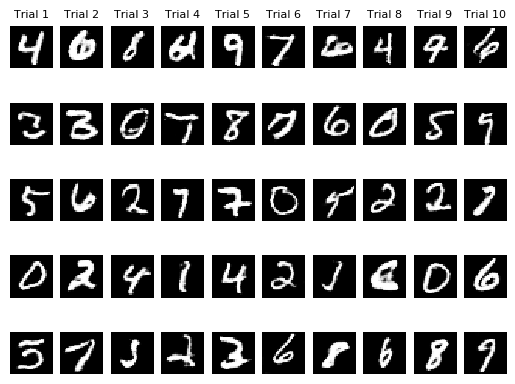}
        \caption{\renyi GAN-GP-$3.0$ sample images.}\label{fig:renyigan_3}
    \end{subfigure}
    \quad
    \begin{subfigure}[t]{0.47\textwidth}
        \centering
        \includegraphics[width=70mm]{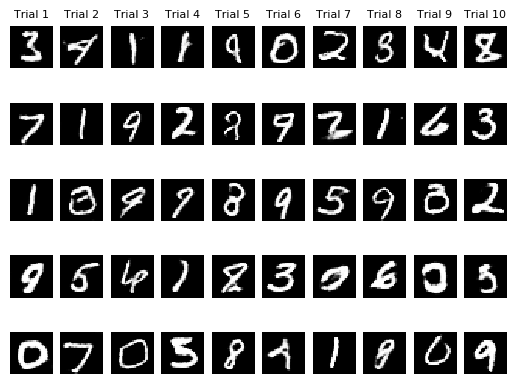}
        \caption{DCGAN-GP-$L_1$ sample images.}\label{fig:dcgan_gp_l1}
    \end{subfigure}
    \quad
    \begin{subfigure}[t]{0.47\textwidth}
        \centering
        \includegraphics[width=70mm]{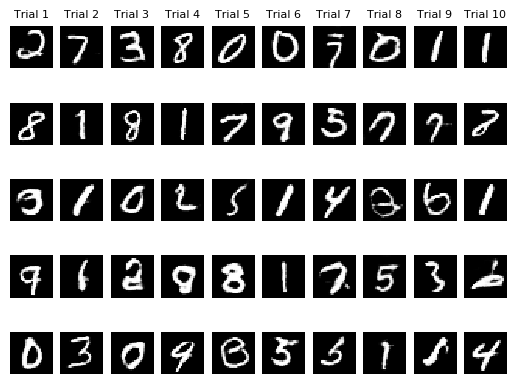}
        \caption{\renyi GAN-GP-$3.0$-$L_1$ sample images.}\label{fig:renyigan_4}
    \end{subfigure}
    \quad
    \caption{MNIST generated images of the best performing \renyi GANs and DCGANS in terms of FID scores.}
    \label{fig:MNIST renyigan generated images}
\end{figure}

A total of ten trials were run while controlling the random seed in each trial.
We did not use the Inception network to calculate the FID scores as
it is not trained on classifying handwritten MNIST images.
Instead, the scores were computed using the raw real and fake images under  
multivariate Gaussian distributions.
The average and variance over ten trials of the best FID scores (i.e., lowest score over all epochs), and the average and variance of the epoch when the best FID score is achieved 
are presented for a representative subset of the tested systems in Table~\ref{table:FID average renyigan}.
A few variants of \renyi GANs converged to meaningful results, which we show
in Table \ref{table:renyigan1 FID average}. 
Finally, samples of MNIST generated images are given in Figure~\ref{fig:MNIST renyigan generated images}.



\begin{table}[hb]
\centering
\caption{\renyi GANs MNIST experiments: 
	the average and variance of the best FID scores and the average and variance of the epoch this occurs (over 10 trials).}
	\label{table:FID average renyigan}
\vskip 0.1in
\begin{tabular}{c C N C C}
\hline \hline
 & Average best FID score & Best FID scores variance & Average epoch & Epoch variance \\ \hline
\renyi GAN-$3.0$ & 52.99 & 292.71 & 32.70 & 92.21 \\ 
	\textbf{\renyi GAN-}$\mathbf{[0, 3]}$ & {\bf 41.59} & 693.61 & 95.40 & 9096.64  \\ 
	DCGAN & 59.061 & $\mathbf{2.01 \times 10^{-28}}$ & {\bf 13.00} & {\bf 0.80} \\ \hline 
\renyi GAN-$3.0$-$L_1$ & 1.80 & 2.95 $\times 10^{-3}$ & 86.80 & 4611.16 \\ 
\textbf{\renyi GAN-$\mathbf{[0, 3]}$-$\mathbf{L_1}$} & \textbf{1.77} & 4.90 $\times 10^{-3}$ & \textbf{36.10} & \textbf{36.80}  \\ 
DCGAN-$L_1$ & 1.93 & 3.83 $\times 10^{-3}$ & 52.30 & 2605.61 \\ 
\hline
\renyi GAN-GP-$3.0$ & \textbf{1.36} & 2.51 $\times 10^{-3}$ & 209.10 & 751.09 \\ 
\renyi GAN-GP-$[0, 3]$ & 1.41 & 3.09 $\times 10^{-3}$ & 201.90 & 1209.69  \\ 
	DCGAN-GP & \textbf{1.36} & {$\mathbf{1.45 \times 10^{-3}}$} & 225.20 & \textbf{342.56} \\ \hline
\textbf{\renyi GAN-GP-}$\mathbf{3.0}$-$\mathbf{L_1}$ & \textbf{1.17} & $ 3.62 \times 10^{-3}$ & 221.70 & \textbf{609.61} \\ 
\renyi GAN-GP-$[0, 3]$-$L_1$ & 1.22 & 6.33 $\times 10^{-3}$ & 224.10 & 1075.09 \\ 
	DCGAN-GP-$L_1$ & 1.18 & 1.58 $\times 10^{-3}$ & {\bf 200.50} & 1263.05 \\ \hline 
\end{tabular}
\end{table}
\begin{table}[htb]
        \centering
        \caption{The average and variance best FID scores for \renyi GANs that generated meaningful MNIST images and
                the average and variance of the epoch when this occurs.}
        \label{table:renyigan1 FID average}
        \vskip 0.1in
        \begin{tabular}{l C N c c}
                \hline \hline
                & Average best FID score & Best FID scores variance & Average epoch & Epoch variance \\ \hline
		\renyi GAN-3.0 & 1.66 & {\bf 0.00} & {\bf 60.00} & {\bf 0.00} \\
		{\bf \renyi GAN-[0,3]} & {\bf 1.36} & 3.81 ${\times 10^{-3}}$ & 240.00 & 6.00 \\
        \hline
        \end{tabular}
\end{table}

\subsubsection{CelebA results}
\begin{figure}[htb]
    \centering
        \begin{subfigure}[t]{.47\textwidth}
         \centering
        \includegraphics[width=0.98\linewidth]{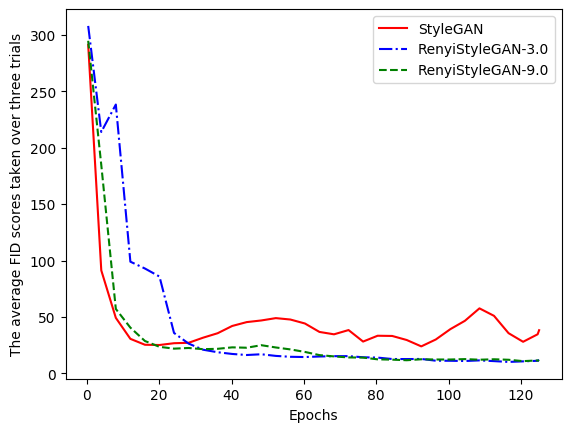}
        \caption{Average FID vs epochs, \renyi StyleGANs and StyleGANs for CelebA ($64 \times 64 $).}
        \label{fig:stylegan_ave_scores}
     \end{subfigure}
	\quad
	\begin{subfigure}[t]{0.47\textwidth}
         \centering
             \includegraphics[width=0.98\linewidth]{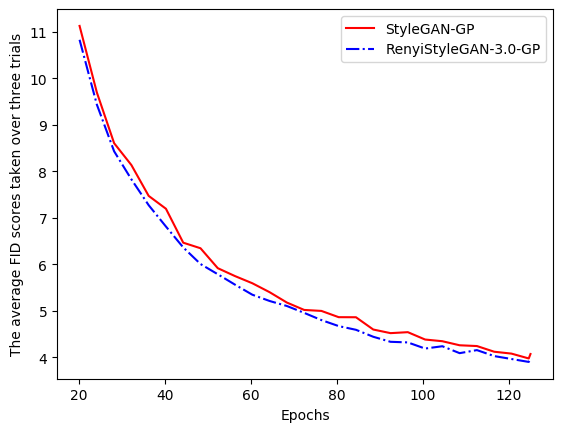}
        \caption{Average FID vs epochs, \renyi StyleGAN-GPs and StyleGAN-GPs for CelebA ($64 \times 64 $).}
        \label{fig:stylegan-gp_ave_scores}
     \end{subfigure}
     \quad
     	\begin{subfigure}[t]{0.47\textwidth}
     	\centering 
             \includegraphics[width=0.98\linewidth]{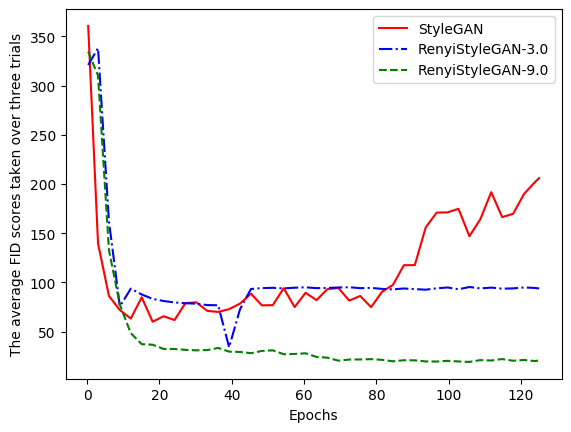}
    \quad
	\caption{Average FID vs epochs, \renyi StyleGANs and StyleGANs for CelebA ($128 \times 128$).}
	\label{fig:renyistylegan128 scores ave}
     \end{subfigure}
    \caption{Evolution of the average FID scores throughout training for \renyi StyleGANs.}
    \label{fig:renyistylegan average plots}
\end{figure}

\begin{figure}[htb]
    \centering
    \begin{subfigure}[t]{0.47\textwidth}
        \centering
        \includegraphics[width=70mm]{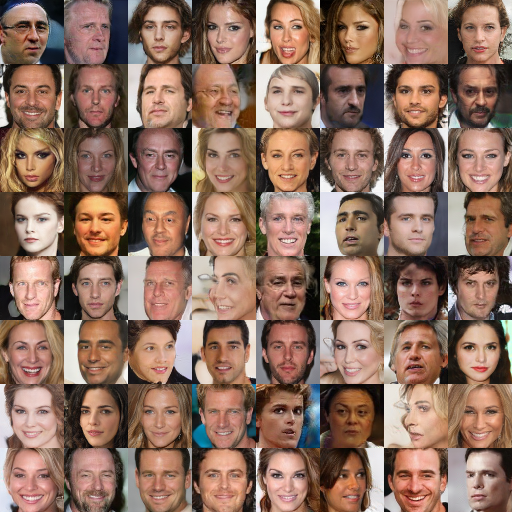}
        \caption{StyleGAN: FID score~16.20.}\label{fig:stylegan_appendix_t1}
    \end{subfigure}
    \quad
    \begin{subfigure}[t]{0.47\textwidth}
         \centering
         \includegraphics[width=70mm]{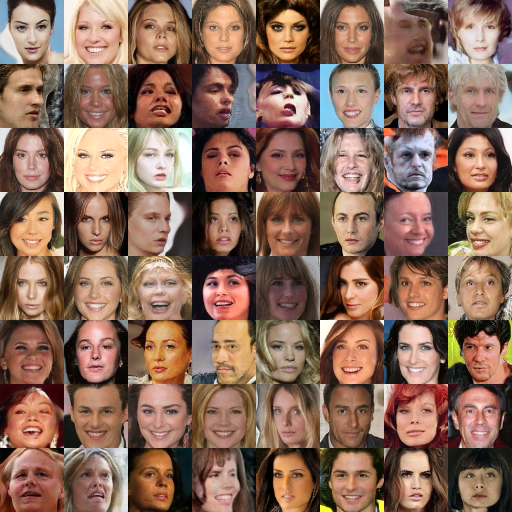}
         \caption{\renyi StyleGAN-3.0: FID score~9.67.}\label{fig:renyistylegan3.0_appendix_t1}
     \end{subfigure}
     \quad
    \begin{subfigure}[t]{0.47\textwidth}
        \centering
        \includegraphics[width=70mm]{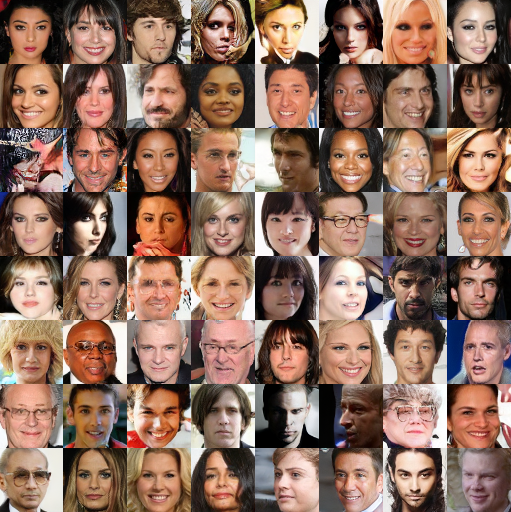}
        \caption{StyleGAN-GP: FID score~4.06.}\label{fig:stylegan_appendixGP_t1}
    \end{subfigure}
    \quad
    \begin{subfigure}[t]{0.47\textwidth}
        \centering
        \includegraphics[width=70mm]{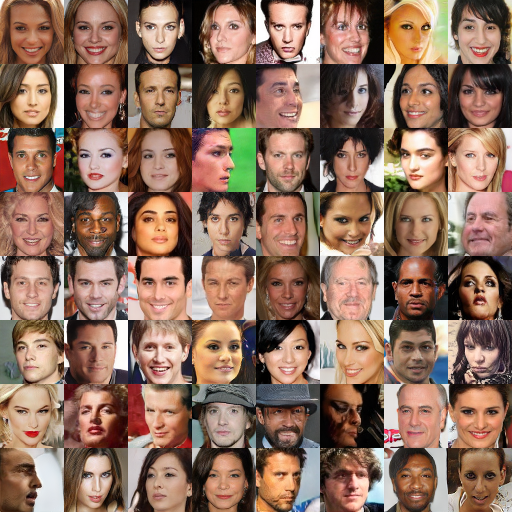}
        \caption{\renyi StyleGAN-3.0-GP: FID score~3.91.}\label{fig:renyistylegan3.0_appendixGP_t1}
    \end{subfigure}
    \quad
       \caption{CelebA (64~$\times$~64) generated images of the best performing \renyi StyleGANs
          for each version and their StyleGAN counterparts for a sample trial.}
    \label{fig:renyistylegan generated images trial 1}
\end{figure}

We demonstrate our results on both $64 \times 64 \times 3$ and 
$128 \times 128 \times 3$ CelebA images.
For the $64 \times 64 \times 3$ case, 
three trials were run for a few $\alpha$ values for \renyi StyleGANs. 
\renyi StyleGANs with and without the simplified gradient penalty 
were also tested. 
Only \renyi StyleGAN-3.0-GP was tested because the addition of the simplified 
gradient penalties increased computing time. 
The FID scores were calculated every $80,000$ images.
The average and variance of best FID scores taken over the three trials are presented in Table \ref{table:FID average renyistylegan}.
We present the plots of the average FID scores taken over the three trials
versus epochs in Figure \ref{fig:stylegan_ave_scores} and~\ref{fig:stylegan-gp_ave_scores}.
Sample generated images of one of the trials for \renyi StyleGAN and StyleGAN are shown in Figure~\ref{fig:renyistylegan generated images trial 1}.

\begin{figure}[htb]
    \centering
    \begin{subfigure}[t]{0.47\textwidth}
        \centering
        \includegraphics[width=70mm]{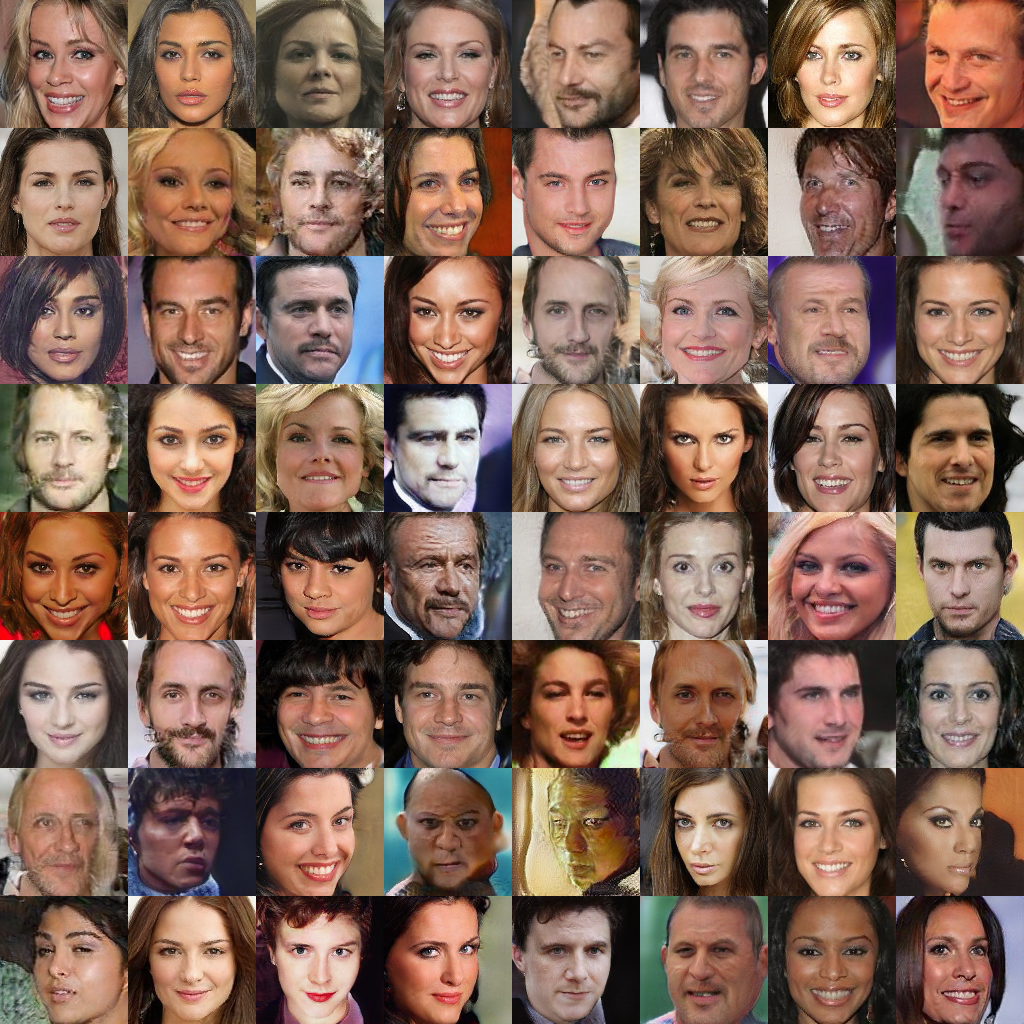}
        \caption{StyleGAN: FID score~19.72.}\label{fig:stylegan128_appendix_t3}
    \end{subfigure}
    \quad
    \begin{subfigure}[t]{0.47\textwidth}
         \centering
         \includegraphics[width=70mm]{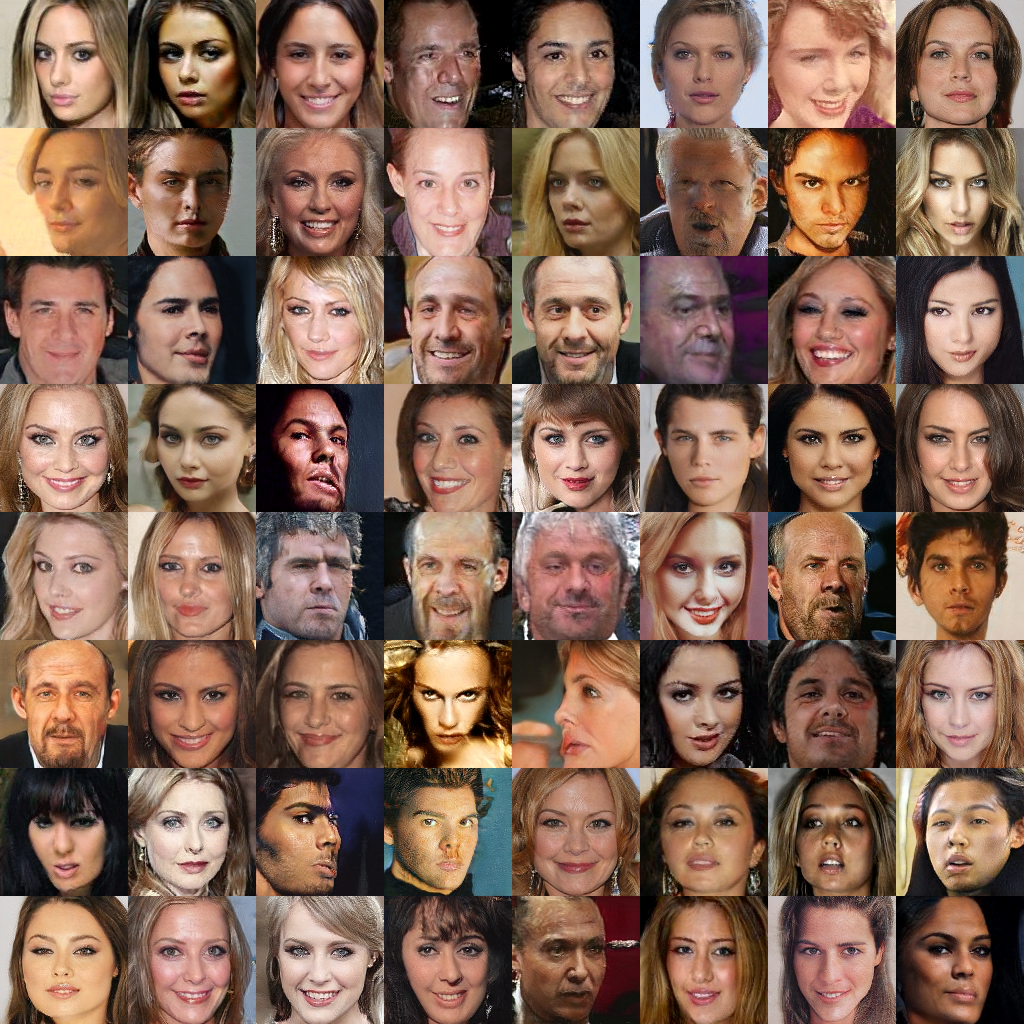}
         \caption{\renyi StyleGAN-3.0: FID score~15.05.}\label{fig:renyistylegan128_3.0_appendix_t3}
     \end{subfigure}
     \quad
    \begin{subfigure}[t]{0.47\textwidth}
        \centering
        \includegraphics[width=70mm]{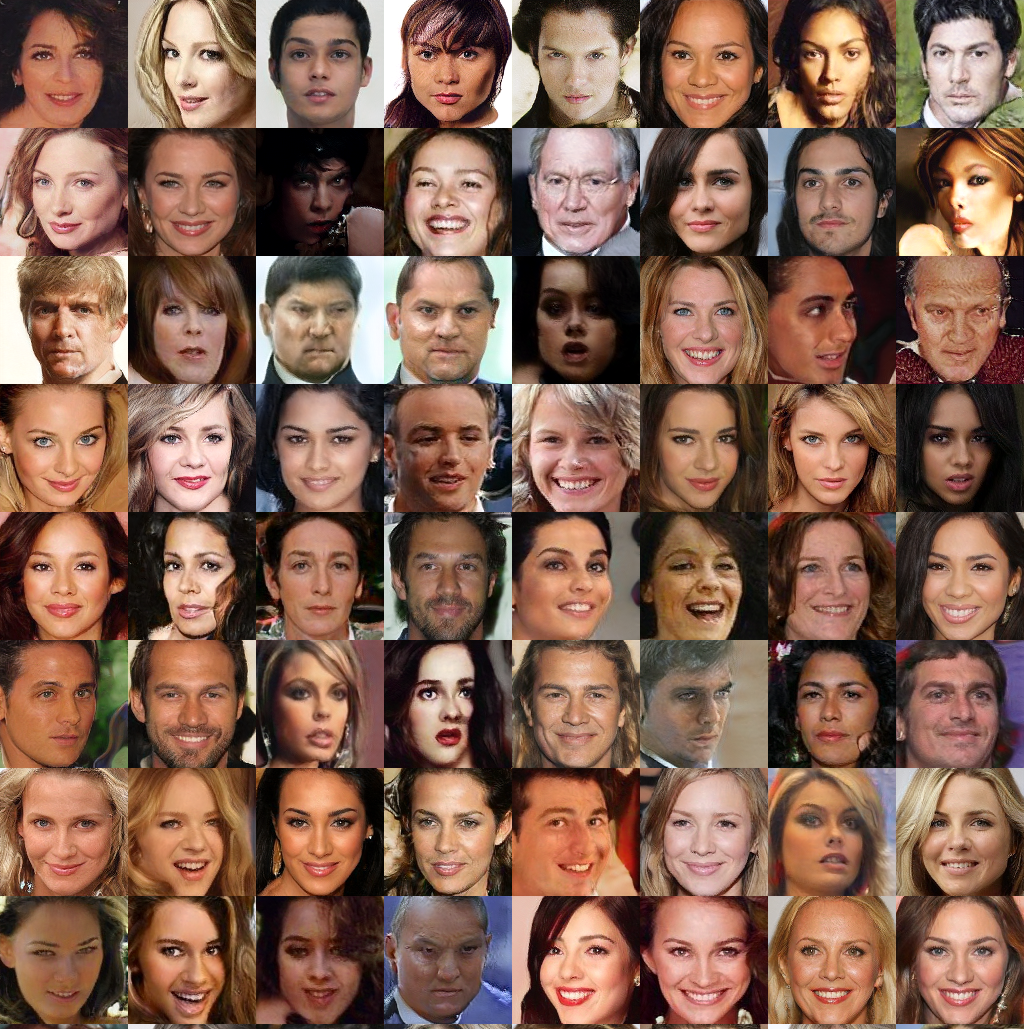}
        \caption{\renyi StyleGAN-9.0: FID score~15.41.}\label{fig:renyistylegan128_9.0_appendix_t3}
    \end{subfigure}
    \quad
       \caption{CelebA (128 $\times$ 128) generated images of the best performing  \renyi Style-GANs
          for each version and their StyleGAN counterparts for a sample trial.}
    \label{fig:renyistylegan128 generated images trial 3}
\end{figure}

For the $128 \times 128 \times 3$ case, 
three trials were also run 
for \renyi StyleGAN and 
StyleGAN with the 
FID scores calculated every $60,000$ images.
The average and variance of best FID scores taken over the three trials are shown in Table \ref{table:FID average renyistylegan128}.
The average FID scores taken over the three trials are plotted 
versus epochs in Figure \ref{fig:renyistylegan128 scores ave}.
Sample generated images of one of the trials for \renyi StyleGAN and StyleGAN are given in Figure~\ref{fig:renyistylegan128 generated images trial 3}.
 
%

\begin{table}[htb]
	\caption{\renyi StyleGANs CelebA (64 $\times$ 64) experiments: the average and variance of the best FID score 
		and the average and variance this occurs (over 3 trials).}
	\vskip 0.1in
	\centering
	\begin{tabular}{L C N C N}
	\hline \hline
        	& Average best FID score & Best FID score variance & Average epoch & Epoch variance \\
	\hline
		{\bf \renyi StyleGAN-3.0} & {\bf 9.80} & $\mathbf{5.93 \times 10^{-2}}$ & 112.54 & {\bf 75.38}  \\
		\renyi StyleGAN-9.0 & 10.38 & 2.12 & 115.22 & 111.28   \\
		StyleGAN & 14.60 & 12.48 & {\bf 93.78} & 477.53 \\ \hline
		{\bf \renyi StyleGAN-3.0-GP} & {\bf 3.88} & $\mathbf{1.84 \times 10^{-3}}$ & 122.05 & {\bf 4.35}  \\
		StyleGAN-GP & 3.92 & 9.06 $\times 10^{-3}$ & {\bf 119.37} & 58.94  \\
	\hline 
	\end{tabular}
	\label{table:FID average renyistylegan}
\end{table}

\begin{table}[hb]
	\caption{\renyi StyleGANs CelebA (128 $\times$ 128) experiments: the average and variance of the best FID score 
		and the average and variance this occurs (over 3 trials).}
	\vskip 0.1in
	\centering
	\begin{tabular}{L C N C N}
	\hline \hline
        	& Average best FID score & Best FID score variance & Average epoch & Epoch variance \\
	\hline
		\renyi StyleGAN-3.0 & 30.47 & 480.69 & 72.50 & 565.68  \\
		{\bf \renyi StyleGAN-9.0} & {\bf 17.34} & {\bf 3.40} & 93.64 & {\bf 371.08}   \\
		StyleGAN & 39.11 & 962.46 & {\bf 48.33} & 918.63 \\ 
	\hline 
	\end{tabular}
	\label{table:FID average renyistylegan128}
\end{table}

\clearpage

\subsection{Discussion}
\subsubsection{MNIST}
The DCGAN baseline exhibited unstable training as was expected 
and the addition of the \renyi \ loss was able to ameliorate convergence but had similar instabilities.
More specifically, \renyi GAN-$[0, 3]$ converged in three out of ten trials, 
achieving an average best FID scores of $1.36$, while DCGAN experienced mode collapse in all ten trials.
This FID score is comparable to that of applying simplified gradient penalty to the network. 
Note that these networks took an average time of~42.33 minutes to train for one trial. 


Applying the $L_1$ normalization drastically improved the convergence of all networks with no computational overhead.
In fact, on average over $250$ epochs and $10$ trials, adding $L_1$ normalization on average decreased the training time for one trial to~41.79 minutes,
which is a decrease of $1.27\%$.
This is expected as the $L_1$ normalization is similar to the L$k$GANs generator loss function when $k = 1$ and $\gamma = -\log(2)$,
which, as we have shown in Section~\ref{sec:lkgan}, improves training stability. 
Using $L_1$ normalization also has the added benefit of networks converging to an optimal FID value in fewer epochs than any other convergent 
networks across all versions. 
We note that \renyi GAN-$[0, 3]$-$L_1$ outperformed all other loss functions in Version~2
and it was sufficient to train it within $50$ epochs.
The development of a rigorous theory that describes this phenomenon is an interesting future direction to better understand the dynamics of GANs.

In Version~$3$, \renyi GAN-GP-$[1.1, 4]$ performed among the best compared to other \renyi GAN-GP variants, with an identical performance to DCGAN-GP.
Moreover, on average it converged to its best FID score in fewer epochs than DCGAN-GP.
Note, however, that the use of gradient penalty increased the average computation time to~47.54 minutes for a single trial, an increase of~$12.30\%$ compared
to Version~1.
The best performing network in terms of FID score was \renyi GAN-GP-$3.0$-$L_1$, seen in the Version~4 results of Table~\ref{table:FID average renyigan}.
Note that \renyi GAN-GP-$0.5$-$L_1$, \renyi GAN-GP-$3.0$-$L_1$, \renyi GAN-GP-$[0, 0.9]$-$L_1$, and DCGAN-GP-$L_1$ 
exhibited quite similar FID scores. 
However, on average, DCGAN-GP-$L_1$ converged to its optimal FID score in fewer epochs than its counterparts.
On average, these networks with gradient penalty and $L_1$ normalization took~47.17 minutes to train for a single trial, 
which is a slight decrease in computational time compared to applying simplified gradient penalties only. 


In summary, the extra degree of freedom provided by the order $\alpha$ resulted in equivalent or better FID scores in fewer epochs when using either $L_1$ normalization or
the simplified gradient penalty.
The greatest advantage of \renyi GANs when applied to MNIST was its ability to 
converge to realistic and diverse generated images quicker than DCGANs in most versions.

\subsubsection{CelebA}
For $64 \times 64 \times 3$ CelebA, we observed that \renyi StyleGANs (with $\alpha>1$) outperform StyleGANs 
in terms of FID scores, with setting $\alpha = 3.0$ achieving the best average FID score.
However, further investigation on the best range of values of $\alpha$ for \renyi StyleGANs is necessary.

Comparing the performance of \renyi StyleGAN to StyleGAN in Figure~\ref{fig:stylegan_ave_scores} reveals that \renyi StyleGAN performs consistently and does not display the erratic unstable behaviour of regular StyleGANs.
One explanation for the difference in performance dynamics is that the \renyi \ loss dampens the loss of each individual 
sample in the batch, reducing the effect of samples that may be given spurious gradient directions.
Combined with our use of the Adam optimizer 
to keep track of the gradient variance, 
the overall effect that dampening has on the entire objective 
function is normalized out, while still maintaining the benefit of dampening individual samples from the generator.
Additionally, as we close the gap between StyleGANs with and without gradient penalty, one benefit of not needing gradient penalty 
was the significant reduction in computation time: 
\renyi StyleGAN-$3.0$ took~25.63 hours without gradient penalty 
and 31.8 hours with gradient penalty, 
yielding a 24$\%$ increase in computation time when using gradient penalty. 
Lastly, both \renyi StyleGAN-GP and StyleGAN-GP performed identically;
see Figure~\ref{fig:stylegan-gp_ave_scores}.


For the $128 \times 128 \times 3$ CelebA dataset, we observed that StyleGAN converged to meaningful results in two out the three trials, whereas both \renyi StyleGAN-3.0 and \renyi StyleGAN-9.0 converged to meaningful results in all three trials. 
Furthermore, Figure~\ref{fig:renyistylegan128 scores ave} confirms that for a large $\alpha$ value, \renyi StyleGANs does not behave erratically during training. 
Although StyleGAN generated high quality images early in training, it suffered from mode collapse as training continued. 
In contrast, both \renyi StyleGANs typically generated high quality images as training continued, though the FID of one \renyi StyleGAN-3.0 run increased as training continued and consequently dragged the average FID up in Figure \ref{fig:renyistylegan128 scores ave}. 
This behaviour is useful as it bridges the gap between StyleGANs with and without gradient penalty.



\subsection{Comparing \renyi GANs and L$k$GANs}
We observe that the quality of generated images in terms of FID scores was better for \renyi GANs than for L$k$GANs, irrespective of the tested datasets. 
For the $64 \times 64 \times 3$ CelebA dataset, the best performing L$k$StyleGAN had parameters $k = 1.0$ with $a=0$, $b=1$, and $c=1$, which realized an FID score of~$18.83$.
Although it outperformed its LSStyleGAN counterpart, L$k$StyleGAN achieved a higher average FID score than \renyi StyleGAN. 
Indeed, \renyi StyleGAN with $\alpha = 3.0$ and $\alpha = 9.0$ had FID scores of~$9.80$ and~$10.38$, respectively.
The use of gradient penalty showed a similar pattern; \renyi StyleGAN-$3.0$ had a lower average FID score 
compared to the best performing L$k$StyleGAN (3.88 vs.~4.31).
Similarly, for the MNIST dataset, the best performing L$k$GAN achieved an average FID score of~$3.13$ (not shown here, see Footnote~\ref{mnist-footnote}), whereas the best performing \renyi GAN achieved an average FID score of~$1.77$. 
Hence experiments show that \renyi GANs consistently outperformed L$k$GANs in terms of generated image quality. 

Furthermore, \renyi GANs outperformed L$k$GANs in terms of training stability. 
For the CelebA dataset, \renyi StyleGANs did not exhibit the erratic fluctuations in image quality of L$k$StyleGANs during training. 
This is evident when comparing the plots of average FID score versus epochs for L$k$StyleGANs and \renyi StyleGANs in Figures~\ref{fig:lkstylegan average plots} and~\ref{fig:renyistylegan average plots}, respectively.
Indeed \renyi GANs' increased training stability bridges the gap between GANs with and without gradient penalty. 
Finally, under gradient penalty, both L$k$StyleGANs and \renyi StyleGANs performed similarly. 


%% file: conclusion.tex
\section{Conclusion}
\label{sec:conclusion}
We introduced two new GAN generator loss functions. 
We first analyzed and implemented L$k$GANs, which generalize LSGANs (for $k\neq 2$). 
We showed that the theoretical minimum when solving the generator optimization problem for L$k$GANs is achieved when the generator's distribution matches the true data distribution. 
Using experiments on the 
CelebA dataset under the StyleGAN architecture, the new L$k$GANs loss functions conferred greater training stability and better generated image quality than LSGANs. 
Experiments also revealed new research directions, such as analyzing the effects of the L$k$GAN parameters and the order $k$ on performance.


We next proposed, analyzed, and implemented a GAN generator loss function based on \renyi \ cross-entropy measures of order $\alpha$ ($\alpha>0$ and $\alpha \neq 1$). 
We showed that the classical GANs analytical minimax result expressed in terms of minimizing the Jensen-Shannon divergence between the generator and the unconstrained discriminator distributions can be generalized for any $\alpha$ in terms of the broader Jensen-\renyi \ divergence, with the original GANs loss function provably recovered in the limit of $\alpha$ approaching 1. 
We demonstrated via experiments on MNIST and CelebA datasets that the \renyi-based loss function (which was analytically shown to be stable for $\alpha\ge 2$) yields performance improvements over the original GANs loss function in terms of the quality of the generated images and training stability.\footnote{Similar performance and stability results for both L$k$StyleGAN and \renyi StyleGAN were also observed in experiments (not reported here) conducted on the CIFAR10 dataset.}
In particular, we observed that \renyi GANs used with $L_1$ normalization on MNIST images does not need gradient penalty to reduce the GAN mode collapse problem. 
Furthermore, \renyi StyleGANs (used on CelebA images) provide more robust convergence dynamics than StyleGANs and can dispose of using gradient penalty without affecting image fidelity while requiring considerably less training time.
Overall, the performance 
of \renyi GAN/\renyi StyleGAN was superior to those of L$k$GAN/L$k$StyleGAN by virtue of the  \renyi \ cross-entropy loss function.
For future research, it would be useful to examine the effects of the \renyi \ order $\alpha$ on the optimal choice of network architectures and parameters. 
Finally, we note that the L$k$- and \renyi-centric methods studied in this work can be judiciously adopted to other deep learning neural network architectures.



%% file: appendix.tex
\section{Neural network architectures} 
\label{architectures}
The StyleGAN architecture were taken directly from \cite{styleGAN}.
The architectures for the MNIST dataset are detailed below in Tables \ref{genArch} and \ref{disArch}. 
We used the architecture guidelines provided by \cite{Radford2015} and the GANs tutorial in \cite{tensorflow}. 
We shorten some of the common terms used to describe the layers of the networks.
A fully connected layer in a neural network is denoted by FC, while we have used upconv. to denote a convolution layer that is
specifically padded to increase the dimensions of its input image.
The bias in each upconv. layer was not used in an effort to reduce the amount of parameters and the computational training time. 
The parameters of the neural networks were initialized by sampling a Gaussian random variable with mean $0$ and standard deviation
$0.01$.

\begin{table}[H]
\caption{The generator's architecture for MNIST dataset.}
\label{genArch}
\vskip 0.15in
\begin{center}
\begin{small}
\begin{sc}
\begin{tabular}{l}
\hline \hline
Generator \\
\hline
	Input multivariate Gaussian noise vector of size $784$ with mean $\mathbf{0}$ \\ 
	\ \ \ and 
	a covariance matrix that is the identity matrix of size $784 \times 784$. \\
	FC to $12,544$ neurons. \\
        Reshape into $7 \times 7 \times 256$ image.\\
        $5 \times 5$ upconv. $128$ LeakyRELU, batchnorm. \\
        $5 \times 5$ upconv. $64$ LeakyRELU, stride 2, batchnorm. \\
        $5 \times 5$ upconv. $1$ channel, $\tanh$ activation. \\
\hline
\end{tabular}
\end{sc}
\end{small}
\end{center}
\vskip -0.1in
\end{table}

\begin{table}[H]
\caption{The discriminator's architecture for MNIST dataset.}
\label{disArch}
\vskip 0.15in
\begin{center}
\begin{small}
\begin{sc}
\begin{tabular}{l}
\hline \hline
Discriminator \\
\hline
        Input $28 \times 28 \times 1$ grey image. \\
        $5 \times 5$ conv. $64$ LeakyRELU, stride 2, batchnorm, \\
        \ \ \ \ dropout $0.3$. \\
        $5 \times 5$ conv. $128$ LeakyRELU, stride 2, batchnorm, \\
        \ \ \ \ dropout $0.3$. \\
	FC to one output, $\frac{1}{1 + e^{-x}}$ activation. \\
\hline
\end{tabular}
\end{sc}
\end{small}
\end{center}
\vskip -0.1in
\end{table}

\section{Algorithms}
\label{algorithms}
We present the algorithms for L$k$GANs below.
For the CelebA dataset, the constants of the algorithms are $n = 125$ epochs or 25 million images and the batch size $m = 126$.

We also present the algorithms for \renyi GANs.
For the MNIST dataset, the constants of the algorithms are $n = 100$ epochs or 6 million images and batch size $m = 100$.
For the CelebA dataset, the constants of the algorithms are $n = 125$ epochs or 25 million images and the batch size $m = 126$.

\begin{algorithm}[htb]
	\caption{Overview of L$k$GAN-v$1$-$k$ and L$k$GAN-v$2$-$k$ algorithms with and without the simplified gradient penalty.}
	\label{algo:lkgan}
\begin{algorithmic}[tb]
	\STATE {\bfseries Initialize} neural networks.
        \STATE {\bfseries Fix} number of epochs $n$.
        \FOR{$i = 0$ {\bfseries to } $n-1$}
		\STATE {\bfseries Sample} batch size of $m$ noise samples $\{\bm{z}_1, \ldots, \bm{z}_m\}$ from noise prior $p_\mathbf{Z}$
		\STATE {\bfseries Sample} batch size of $m$ examples $\{\mathbf{x}_1, \ldots, \mathbf{x}_m \}$ from the true distribution $p_\mathbf{X}$
                \STATE {\bfseries Update} the discriminator by descending its gradient without the simplified gradient penalty:
                        $$\nabla_{\bm{\tilde{\theta}}} \left(\frac{1}{m}\sum_{i = 1}^{m} \left[
				\frac{1}{2} (D(\mathbf{x}_i) - b)^2 + \frac{1}{2} (D(g(\bm{z}_i)) - a)^2
				\right]\right),$$
			or with the simplified gradient penalty:
			\begin{eqnarray*}
				&& \nabla_{\bm{\tilde{\theta}}} \Bigg(\frac{1}{m}\sum_{i = 1}^{m} \left[
					\frac{1}{2} (D(\mathbf{x}_i) - b)^2 + \frac{1}{2} (D(g(\bm{z}_i)) - a)^2
                                \right] \\
			&& + 5 \left ( \frac{1}{m} \sum_{i=1}^{m} \left
              				\Vert \nabla_{\mathbf{x}} 
				\log\left(\frac{D(\mathbf{x})}{1 - D(\mathbf{x})} \right) \bigg \vert_{\mathbf{x}= \mathbf{x}_i}  \right \Vert_2 ^2  \right) \Bigg).
			\end{eqnarray*}
		\STATE{Update} the generator by descending its gradient:
                $$ \nabla_{\bm{\theta}}   \left(
			\frac{1}{m} \sum_{i = 1}^{m} \vert D(g(\bm{z}_i)) - c \vert^{k} \right),  $$
        \ENDFOR
\end{algorithmic}
\end{algorithm}


\begin{algorithm}[htb]
	\caption{Overview of \renyi GAN-$\alpha$, \renyi GAN-$\alpha$-$L_1$, and \renyi GAN-GP-$\alpha$-$L_1$ algorithms}
	\label{algo:renyigan}
\begin{algorithmic}[tb]
	\STATE {\bfseries Initialize} neural networks. 
        \STATE {\bfseries Fix} number of epochs $n$.
        \FOR{$i = 0$ {\bfseries to } $n-1$}
                \STATE {\bfseries Sample} batch size of $m$ noise samples $\{\bm{z}_1, \ldots, \bm{z}_m\}$ from noise prior $p_{\mathbf{Z}}$
                \STATE {\bfseries Sample} batch size of $m$ examples $\{\mathbf{x}_1, \ldots, \mathbf{x}_m \}$ from the true distribution $p_{\mathbf{X}}$
                \STATE {\bfseries Update} the discriminator by descending its stochastic gradient without the simplified gradient penalty:
                        $$\nabla_{\bm{\tilde{\theta}}} \bigg(-\frac{1}{m}\sum_{i = 1}^{m} \big[
                                        \log D(\mathbf{x}_i)
                                        + \log(1 - D(g(\bm{z}_i)))
                                \big]\bigg),$$
                or with the simplified gradient penalty:
		\begin{eqnarray*}
			&&\nabla_{\bm{\tilde{\theta}}} \Bigg(-\frac{1}{m}\sum_{i = 1}^{m} \big[
                                        \log D(\mathbf{x}_i)
                                        + \log(1 - D(g(\bm{z}_i)))
				\big]  \\ && +  
			5 \left ( \frac{1}{m} \sum_{i=1}^{m} \left \Vert \nabla_{\mathbf{x}} 
			\log\left(\frac{D(\mathbf{x})}{1 - D(\mathbf{x})} \right) \bigg \vert_{\mathbf{x}= \mathbf{x}_i}  \right \Vert_2 ^2  \right) \Bigg),
		\end{eqnarray*}
                and update the generator by descending its stochastic gradient without $L_1$ normalization:
                $$ \nabla_{\bm{\theta}}  \frac{1}{\alpha - 1}\log \Bigg[ \left(
	\frac{1}{m} \sum_{i = 1}^{m} [1 - D(g(\bm{z}_i))]^{\alpha - 1} \right) \Bigg], $$
                or with $L_1$ normalization:
$$ \nabla_{\bm{\theta}} \Bigg \vert \frac{1}{\alpha - 1}\log \Bigg[ \left(
	\frac{1}{m} \sum_{i = 1}^{m} [1 - D(g(\bm{z}_i))]^{\alpha - 1} \right) \Bigg] + \log(2) \Bigg \vert,$$
        \ENDFOR
\end{algorithmic}
\end{algorithm}